\documentclass{scrartcl}
\usepackage{amsmath}
\usepackage{amsfonts}
\usepackage{amssymb}
\usepackage{dsfont}
\usepackage[T1]{fontenc}
\usepackage[latin1]{inputenc}
\usepackage{epsfig}
\usepackage{graphicx}

\usepackage[boxed, lined]{algorithm2e}
\usepackage{float}
\usepackage{comment}
\usepackage{mathtools}

\newcommand{\ba}{\mathbf{a}}
\newcommand{\bb}{\mathbf{b}}

\newcommand{\bx}{\mathbf{x}}

\newcommand{\bz}{\mathbf{z}}
\newcommand{\bw}{\mathbf{w}}
\newcommand{\bv}{\mathbf{v}}

\newcommand{\bX}{\mathbf{X}}
\newcommand{\D}{{\mathcal{D}}}

\newcommand{\W}{{\mathcal{W}}}

\newcommand{\Nu}{{\mathcal{N}}}

\newcommand{\N}{\mathbb{N}}

\newcommand{\R}{\mathbb{R}}
\newcommand{\Z}{\mathbb{Z}}
\newcommand{\Rd}{\mathbb{R}^d}

\newcommand{\balpha}{\mathbf{\alpha}}

\newcommand{\beq}{\begin{eqnarray*}}

\newcommand{\eeq}{\end{eqnarray*}}

\newcommand{\beqm}{\begin{eqnarray}}

\newcommand{\eeqm}{\end{eqnarray}}

\newtheorem{theorem}{Theorem}

\newtheorem{lemma}{Lemma}
\newtheorem{definition}{Definition}

\newcommand{\EXP}{{\mathbf E}}
\newcommand{\PROB}{{\mathbf P}}

\renewcommand{\P}{{\cal P}}
\DeclareOldFontCommand{\bf}{\normalfont\bfseries}{\mathbf}
\DeclareOldFontCommand{\it}{\normalfont\itshape}{\mathbf}

\allowdisplaybreaks
\begin{document}
\renewcommand{\thefootnote}{\fnsymbol{footnote}}
\newcommand{\F}{{\cal F}}
\newcommand{\Sp}{{\cal S}}
\newcommand{\G}{{\cal G}}
\newcommand{\HH}{{\cal H}}

\begin{center}

  {\LARGE \bf
    Analysis of the rate of convergence of an over-parametrized
convolutional neural network image classifier learned by gradient descent
  }
\footnote{
Running title: {\it Over-parametrized convolutional neural networks}}
\vspace{0.5cm}

Michael Kohler$^{1}$,
Adam Krzy\.zak$^{2,}$\footnote{Corresponding author. Tel:
  +1-514-848-2424 ext. 3007, Fax:+1-514-848-2830}
and Benjamin Walter$^{1}$

{\it $^1$
Fachbereich Mathematik, Technische Universit\"at Darmstadt,
Schlossgartenstr. 7, 64289 Darmstadt, Germany,
email: kohler@mathematik.tu-darmstadt.de, bwalter@mathematik.tu-darmstadt.de}

{\it $^2$ Department of Computer Science and Software Engineering, Concordia University, 1455 De Maisonneuve Blvd. West, Montreal, Quebec, Canada H3G 1M8, email: krzyzak@cs.concordia.ca}

\end{center}
\vspace{0.5cm}

\begin{center}
March 21, 2023
\end{center}
\vspace{0.5cm}

\noindent
    {\bf Abstract}\\
Image classification based on over-parametrized convolutional
neural networks with a global average-pooling layer is considered.
The weights of the network are learned by gradient descent.
A bound on the rate of convergence of the difference
between the misclassification risk of the newly introduced convolutional
neural network estimate and the minimal possible value is derived.

    \vspace*{0.2cm}


\vspace*{0.2cm}


\section{Introduction}
\label{se1}
\subsection{Scope of this paper}
\label{se1sub1}
In deep learning, the task is to estimate the functional relationship between input and output using deep neural networks. For the particular application area of image classification, the input data consists of observed images and the output data represents classes of the corresponding images that describe what kind of objects are present in the images. The most successful methods, especially in the area of image classification can be attributed to deep learning approaches (see, e.g., Krizhevsky, Sutskever and Hinton (2012), LeCun, Bengio and Hinton (2015), and Rawat and Wang (2017)) and, in particular, to convolutional neural networks (CNNs). Recently, it has been shown that CNN image classifiers that minimize empirical risk are able to achieve dimension reduction (see Kohler, Krzy\.zak and Walter (2022), Kohler and Langer (2021), Walter (2021) and Kohler and Walter (2022)). However, in practice, it is not  possible to compute the empirical risk minimizer. Instead, gradient descent methods are used to obtain a small empirical risk. Furthermore, the network topologies used in practice are over-parameterized, i.e., they have many more trainable parameters than training samples.

The goal of this work is to derive the rate of convergence results for over-parameterized CNN image classifiers, which are trained by gradient descent. Thus this work should provide a better theoretical understanding of the empirical success of CNN image classifiers.
\subsection{Image classification}
\label{se1sub2}
We use the following statistical setting for image classification:
Let
$d_1,d_2 \in \N$ and let
$(\bX,Y)$, $(\bX_1,Y_1)$, \dots, $(\bX_n,Y_n)$
be independent and identically distributed random variables
with values in
\[
[0,1]^{
\{1, \dots, d_1\} \times \{1, \dots, d_2\}
  } \times \{0,1\}.
\]
Here we use the notation
\[
[0,1]^J
= \left\{
(a_j)_{j \in J}
\, : \,
a_j \in [0,1] \quad (j \in J)
\right\}
\]
for a nonempty and finite index set $J$, and we describe a (random)
image from (random) class $Y \in \{0,1\}$ by a (random) matrix $X$
with $d_1$ columns and $d_2$ rows, which contains at position $(i,j)$
the grey scale value of the image pixel at the corresponding position.

Let
\begin{equation}
\label{se1eq2}
\eta(\bx) = \PROB\{ Y=1|\bX=\bx\}
\quad
( \bx \in [0,1]^{
\{1, \dots, d_1\} \times \{1, \dots, d_2\}
  })
\end{equation}
be the so--called a posteriori probability. Then
we have
\[
\min_{f:
  [0,1]^{\{1, \dots, d_1\} \times \{1, \dots, d_2\}}
  \rightarrow \{0,1\}
}
  \PROB\{ f(\bX) \neq Y \}
  =
    \PROB\{ f^*(\bX) \neq Y \},
    \]
    where
\[
f^*(\bx)=
\begin{cases}
  1, & \mbox{if } \eta(\bx) > \frac{1}{2} \\
  0, & \mbox{elsewhere}
  \end{cases}
\]
is the so--called Bayes classifier
(cf., e.g., Theorem 2.1 in Devroye, Gy\"orfi and Lugosi (1996)).
Set
\[
\D_n = \left\{
(\bX_1,Y_1), \dots, (\bX_n,Y_n)
\right\}.
\]
In the sequel we consider the problem of constructing
a classifier
\[
f_n=f_n(\cdot, \D_n):  [0,1]^{\{1, \dots, d_1\} \times \{1, \dots, d_2\}}
\rightarrow \{0,1\}
\]
such that the misclassification risk
\[
\PROB\{ f_n(\bX) \neq Y | \D_n\}
\]
of this classifier is as small as possible. Our aim
is to derive a bound on the expected difference of
the misclassification risk of $f_n$ and the optimal
misclassification risk, i.e., we want to derive an upper bound on
\begin{eqnarray*}
  &&
\EXP\left\{
\PROB\{ f_n(\bX) \neq Y | \D_n\}
-
\min_{f:
  [0,1]^{\{1, \dots, d_1\} \times \{1, \dots, d_2\}}
  \rightarrow \{0,1\}
}
\PROB\{ f(\bX) \neq Y \}
\right\}
\\
&&
  =
\PROB\{ f_n(\bX) \neq Y \}
-
    \PROB\{ f^*(\bX) \neq Y \}.
\end{eqnarray*}

It is well-known that one needs to impose regularity conditions on the
underlying distribution in order to derive non-trivial rate of
convergence results for the error of the misclassification risk of any
estimate in pattern recognition (cf., e.g., Cover (1968) and Devroye and Wagner (1980)).
In the sequel we will assume that our a posteriori probability
satisfies the model introduced below (see Definition 1), which is a  modification of the generalized hierarchical max-pooling model introduced in Kohler, Krzy\.zak and Walter (2022).

The generalized hierarchical max-pooling model, which is also used in slightly modified form in Kohler and Langer (2021), Walter (2021), and in Kohler and Walter (2022), is motivated by the two ideas that, firstly, the object to be classified is contained in a subpart of the image and, secondly, that an image is hierarchically composed of neighboring subparts.
The first idea is realized by looking at each subpart of the image and
estimating for each subpart the probability that it contains the
corresponding object. It is then assumed that the probability for the
entire image corresponds to the maximum of the probabilities of all
subparts of the image. The difference between the previous model and our new model is that instead
of the maximum, we compute an average over all subparts
 (see Definition \ref{de1} a)).
The advantage here is that our new model includes classification tasks
where an image must contain multiple objects at possibly different
image positions, but also classification tasks where only a single object
has to be detected (in case that at each subpart the product of the
probability
that the subpart contains the object and
a constant greater than 1  is estimated).
The second idea realized in the hierarchical max-pooling model is that
the probability for a subpart of the image is hierarchically composed
of decisions of smaller neighboring subparts. This idea is not
realized in our new model introduced below.

\begin{definition}
	\normalfont
	\label{de1}
	Let $d_1,d_2\in\N$ with $d_1,d_2>1$ and $m: [0,1]^{\{1, \dots, d_1\} \times \{1, \dots, d_2\}} \rightarrow \R$.
	
	\noindent
	{\bf a)} Let $\kappa \in \N$ with $\kappa \leq \min\{d_1,d_2\}$
        and set
\[
I=\{0, \dots, \kappa-1\} \times \{0, \dots, \kappa-1\}.
\]
	We say that $m$
	satisfies a {\bf average-pooling model with size $\kappa^2$},
	if there exists a function $f:[0,1]^{(1,1)+I}\rightarrow \R$  such that
	\[
	m(\bx)=
\frac{1}{(d_1 -\kappa+1) \cdot (d_2 - \kappa +1)} \cdot
		\sum_{
			(i,j) \in \Z^2 \, : \,
			(i,j)+I \subseteq \{1, \dots, d_1\} \times \{1, \dots, d_2\}
		}
		f\left(
		\bx_{(i,j)+I}
		\right).
		\]
	
	\noindent
	{\bf b)}
	Let $p \in(0,\infty)$.
	We say that a {\bf average-pooling model of size $\kappa^2$ has smoothness constraint $p$},
	if the function $f$ in the definition of $m$ is $(p,C)$--smooth for some $C>0$ (see Subsection \ref{se1sub5} for the definition of $(p,C)$--smoothness).
\end{definition}

\subsection{Convolutional neural networks}
\label{se1sub3}
The starting point in the construction of our estimate are
convolutional neural networks with $L\in\N$ convolutional layers, one linear layer and one
average-pooling layer for a $[0,1]^{\{1,\dots,d_1\}\times\{1,\dots,d_2\}}$--valued input, where $d_1,d_2\in\N$. These networks have $k_r\in\N$ channels
(also called feature maps)
in the convolutional layer $r$ and the convolution
in layer $r$ is performed
by a window of values of layer $r-1$ of size $M_r\in\{1,\dots,\min\{d_1,d_2\}\}$, where $r \in \{1, \dots, L \}$.
We will denote the input layer as the convolutional layer $0$
with $k_0=1$ channels.
The average-pooling layer will depend on a parameter $M_{L+1}
\in \{1,\dots,\min\{d_1,d_2\}\}$ which describes the size of the window
over which the output of layer $L$ is averaged.

The networks depend on the weight matrix (so--called filter)
\[
\bw
=
\left(
w_{i,j,s_1,s_2}^{(r)}
\right)_{
	1 \leq i,j \leq M_r, s_1 \in \{1, \dots, k_{r-1}\}, s_2 \in \{1, \dots, k_r\},
	r \in \{1, \dots,L \}
},
\]
the weights
\[
\bw_{bias}
=
\left(
w_{s_2}^{(r)}
\right)_{
	s_2 \in \{1, \dots, k_r\},
	r \in \{1, \dots,L\}
}
\]
for the bias in each channel and each convolutional layer and
the output weights
\[
\bw_{out}=(w_{s})_{
	s \in \{1, \dots, k_L\}
}.
\]
For given weight vectors
$\bw$, $\bw_{bias}$ and $\bw_{out}$
the output of the networks is given by a real--valued function on $[0,1]^{\{1,\dots,d_1\}\times\{1,\dots,d_2\}}$ of the form
\begin{eqnarray*}
  &&
	f_{\bw, \bw_{bias}, \bw_{out}}(\bx)\\
&&	=
\frac{1}{
(d_1-M_{L+1}+1) \cdot (d_2-M_{L+1}+1)
}
\cdot
	\sum_{ 	i \in \{1,\dots,d_1-M_{L+1}+1\}, \atop
	j \in \{1,\dots,d_2-M_{L+1}+1\} }
 \left(
	\sum_{s_2=1}^{k_L}
	w_{s_2} \cdot o_{(i,j),s_2}^{(L)}
	\right),
\end{eqnarray*}
where $o_{(i,j),s_2}^{(L)}$ is
the output of the last convolutional layer, which is
recursively defined
as follows:

We start with
\[
o_{(i,j),1 }^{(0)} = x_{i,j}
\quad \mbox{for }
i \in \{1, \dots, d_1\}~\text{and}~j \in \{1, \dots, d_2\}.
\]
Then we define recursively
\begin{equation}
	\label{s2eq1}
	o_{(i,j),s_2}^{(r)}
	=
	\sigma \left(
	\sum_{s_1=1}^{k_{r-1}}
	\sum_{\substack{t_1,t_2 \in \{1, \dots, M_r\}\\(i+t_1-1,j+t_2-1)\in D}}
	w_{t_1,t_2,s_1,s_2}^{(r)}
	\cdot
	o_{(i+t_1-1,j+t_2-1),s_1}^{(r-1)}
	+
	w_{s_2}^{(r)}
	\right)
\end{equation}
for  the index set $D=\{1,\dots,d_1\}\times\{1,\dots,d_2\}$, $(i,j)\in D$, $s_2\in\{1,\dots,k_r\}$
and
$r \in \{1, \dots, L\}$, where $\sigma:\R \rightarrow \R$ is the
activation
function of the convolutional neural network, for which we will
use throughout this paper the logistic squasher
defined by
$\sigma(x)=1/(1+e^{-x})$.

In this paper we consider a special topology of the network where
we compute a huge number of the above convolutional networks in
parallel
and the output of the network is then defined as a linear combination
of the outputs of all those networks. Here all weights (including the
weights used in the linear combination of the networks) will
be learned by gradient descent starting with some proper
(random) initialization, cf. Section \ref{se2} concerning the details.

\subsection{Main result}
\label{se1sub4}
In this paper we introduce an over-parametrized convolutional neural
network image classifier where all weights are learned by gradient
descent. We show that in case that the a posteriori probability
satisfies conditions of an average-pooling model of size $\kappa^2$ and with
smoothness constraint $p \in [1/2,1]$,  a proper random
initialization of our weights together with proper choices
of the stepsize and the number of gradient descent steps
results for any $\epsilon>0$ in
	\begin{eqnarray*}
			&&\PROB\{f_n(\bX) \neq Y\}
			-
			\min_{f: [0,1]^{\{1, \dots, d_1\} \times \{1, \dots, d_2\}} \rightarrow \{0,1\}}
			\PROB\{f(\bX) \neq Y\}
\\
&& \leq
			c_1 \cdot n^{- \frac{1}{2 \cdot \kappa^2 +2}+\epsilon}.
	\end{eqnarray*}
The  upper bound above on the
the difference between the misclassification risk of the newly introduced convolutional
neural network estimate and the minimal possible risk (Bayes risk)  does not depend
on the dimension $(d_1,d_2)$ of the image, which shows that our
convolutional neural network estimate is able to achieve some kind
of dimension reduction. As far as we know the above result is
the first rate of convergence result derived for convolutional
neural network estimates where the weights are learned by
gradient descent (using only one single random initialization).
Our proof relies on the techniques recently developed by Drews and
Kohler (2022) and Kohler and Krzy\.zak (2022) for the analysis
of over-parametrized deep feedforward neural networks learned
by gradient descent and our
main achievement is to demonstrate that these techniques
can also be used to analyze the rates of convergence of
over-parametrized convolutional neural network estimates
learned by gradient descent.

\subsection{Discussion of related results}
\label{se1sub5}
Deep neural networks have been studied intensively in the last decade and applied widely in different domains, see
Goodfellow, Bengio and Courville (2016).
Theoretical analysis of deep network learning has been actively pursued in recent years, see Berner et al. (2021) for a recent survey of progress in mathematics of deep learning. Among different deep network architectures convolutional neural networks introduced by LeCun (1989) are the most popular. They have been
applied in image classification by
Krizhevsky, Sutskever and Hinton (2012) and Kohler, Krzy\.zak and Walter (2022). In the latter paper the authors investigated the rates of CNN image classifiers.


Several recent papers demonstrated theoretically that backpropagation learning works for deep neural
networks. The most popular approach which emerged in this context is so--called landscape approach.
Choromanska et al. (2015) used random matrix theory to
derive a heuristic argument showing that the  risk  of  most  of  the  local  minima  of the
empirical $L_2$ risk $F_n(\bw)$ is  not  much larger  than  the  risk  of  the  global  minimum.
This claim was validated for neural networks with special activation function by, e.g., Arora et al. (2018),
Kawaguchi (2016), and Du and Lee (2018), which have analyzed gradient descent for neural networks with a linear or quadratic
activation function. No good approximation results exist for such neural networks, and
consequently one cannot deduce from these results good rates of convergence for neural network regression estimates.
Du et al. (2018) analyzed gradient descent learning for neural networks
with one hidden layer and Gaussian inputs.
As they used the expected gradient instead of the gradient
in their gradient descent routine,
one cannot apply their results to derive the rate of convergence
for neural network regression estimates learned by the gradient descent.
Liang et al. (2018) applied gradient descent to a modified loss function in classification,
where it is assumed that the data can be interpolated by a neural network.
Neural tangent kernel networks (NTK) were introduced by Jacot, Gabriel and Honger (2020). They showed that in the infinite-width limit case NTK converges to a deterministic limit kernel which stays constant during Gaussian descent training of the random weights initialized  with the Gaussian distributions. These results were extended by Huang, Du and Xu (2020) to orthogonal initialization which was shown to speed up training of fully connected deep networks. Nitanda and Suzuki (2017) obtained global convergence rate for the averaged stochastic gradient descent for over-parametrized shallow neural networks. Braun et al. (2021) showed rate of convergence $1/\sqrt{n}$ (up to a logarithmic factor) for regression functions that have Fourier transforms with polynomially decreasing tails (an assumption slightly stronger than the finite first moment of the Fourier transform assumption of Barron (1993)).

Recently it was shown in several papers, see, e.g.,
Allen-Zhu, Li and Song (2019), Kawaguchi and Huang (2019)
and the literature cited therein, that the gradient descent leads to
a small empirical $L_2$ risk in over-parametrized neural networks.
Here the results in Allen-Zhu, Li and Song (2019) are proven for
the ReLU activation function and neural networks with a polynomial
size in the sample size. The neural networks in
Kawaguchi and Huang (2019) use squashing activation functions and
are much smaller (in fact, they require only a
linear size in the sample size). In contrast to
Allen-Zhu, Li and Song (2019)
there the learning rate is set to zero for all neurons except
for neurons in the output layer
and consequently in different layers of the network
different learning rates are used.
Actually, they compute a linear least squares estimate with the gradient descent, which is not used in practice.
It was shown in Kohler and Krzy\.zak (2021) that any estimate which
interpolates the training data does not generalize well in a sense
that it can, in general, not achieve the optimal minimax rate of convergence
in case of a general design measure. In recent survey paper
Bartlett, Montanari and Rakhlin (2021)
conjectured that over-parametrization allows gradient descent to find interpolating solutions which implicitly impose regularization, and that over-parametrization leads to benign overfitting. For related results involving the truncated Hilbert kernel regression estimate refer to
Belkin, Rakhlin and Tsybakov (2019)
and to Wyner et al. (2017) for the results involving AdaBoost and random forests. Linear regression in overfitting regime has been also considered in
Bartlett, Long and Lugosi (2020). Benign over-parametrization in shallow ReLU networks has been analyzed by Wang and Lin (2021). They showed $L_2$ error rate of $\sqrt{\log n/n}$ for over-parametrized neural network when the number of hidden neurons exceeds  the number of samples. 

Overparametrized deep neural network multivariable regression function estimates have been analyzed in recent papers. Universal consistency of such estimates was shown for over-parameterized standard deep feedforward neural networks learned by gradient descent by Drews and Kohler (2022). This paper was generalized by Kohler and Krzy\.zak (2022), who studied the rates of convergence. The approach used in the present paper is related to these two papers.
In our proof we control the complexity of our over-parametrized convolutional
neural networks by using metric entropy bounds as in
Li, Gu and Ding (2021).
A different approach based on the Rademacher complexity is presented
in Wang and Ma (2022).


\subsection{Notation}
\label{se1sub5}
  The sets of natural numbers, real numbers and nonnegative real numbers
are denoted by $\N$, $\R$ and $\R_+$, respectively. For $z \in \R$, we denote
the smallest integer greater than or equal to $z$ by
$\lceil z \rceil$, and we denote the greatest integer less than
or equal to $z$ by
$\lfloor z \rfloor$.
The Euclidean norm of $x \in \Rd$
is denoted by $\|x\|$, and we set
\[
\|x\|_\infty = \max\{ |x^{(1)}|, \dots, |x^{(d)}| \}
\]
for $x=(x^{(1)},\dots,x^{(d)})^T \in \Rd$.
For $f:\R^d \rightarrow \R$
\[
\|f\|_\infty = \sup_{x \in \R^d} |f(x)|
\]
is its supremum norm.
A function $f:\R^d \rightarrow \R$ is called
$(p,C)$-smooth, if for every $\balpha=(\alpha_1, \dots, \alpha_d) \in
\N_0^d$
with $\sum_{j=1}^d \alpha_j = q$ the partial derivative
$\frac{
	\partial^q f
}{
	\partial x_1^{\alpha_1}
	\dots
	\partial x_d^{\alpha_d}
}$
exists and satisfies
\[
\left|
\frac{
	\partial^q f
}{
	\partial x_1^{\alpha_1}
	\dots
	\partial x_d^{\alpha_d}
}
(\bx)
-
\frac{
	\partial^q f
}{
	\partial x_1^{\alpha_1}
	\dots
	\partial x_d^{\alpha_d}
}
(\bz)
\right|
\leq
C
\cdot
\| \bx-\bz \|^s
\]
for all $\bx,\bz \in \R^d$.%

Let $\F$ be a set of functions $f:\Rd \rightarrow \R$,
let $x_1, \dots, x_n \in \Rd$, set $x_1^n=(x_1,\dots,x_n)$ and let
$p \geq 1$.
A finite collection $f_1, \dots, f_N:\Rd \rightarrow \R$
  is called an $L_p$ $\varepsilon$--cover of $\F$ on $x_1^n$
  if for any $f \in \F$ there exists  $i \in \{1, \dots, N\}$
  such that
  \[
  \left(
  \frac{1}{n} \sum_{k=1}^n |f(x_k)-f_i(x_k)|^p
  \right)^{1/p}< \varepsilon.
  \]
  The $L_p$ $\varepsilon$--covering number of $\F$ on $x_1^n$
  is the  size $N$ of the smallest $L_p$ $\varepsilon$--cover
  of $\F$ on $x_1^n$ and is denoted by $\Nu_p(\varepsilon,\F,x_1^n)$.

For $z \in \R$ and $\beta>0$ we define
$T_\beta z = \max\{-\beta, \min\{\beta,z\}\}$. If $f:\R^d \rightarrow
\R$
is a function and $\F$ is a set of such functions, then we set
$
(T_{\beta} f)(x)=
T_{\beta} \left( f(x) \right)$.

\subsection{Outline}
\label{se1sub6}
The over-parametrized convolutional neural network estimates considered
in this paper are introduced in Section \ref{se2}. The main result
is presented in Section \ref{se3}. Section \ref{se4} contains the proofs.

\section{Definition of the estimate}
\label{se2}

Throughout the paper we let
$\sigma(x)=1/(1+e^{-x})$
be the logistic squasher and
we define the topology of our convolutional neural networks as
follows:
We compute a large number $K_n \in \N$ of the convolutional
neural networks in Subsection \ref{se1sub3} in parallel, where
for simplicity we use for each of these networks $k_L=1$ and
$w_1=1$ (i.e., we skip the linear combination before the
average-pooling),
and we compute a linear combination of the output of these
$K_n$ convolutional neural networks. Here we use again $k_0=1$.

We set
\begin{equation}
\label{se2eq1}
f_{\bw}(\bx) = \sum_{k=1}^{K_n} w_k \cdot f_{\bw_k, \bw_{bias,k}}(\bx)
\end{equation}
where for $k \in \{1, \dots, K_n\}$
\begin{equation}
\label{se2eq2}
f_{\bw_k, \bw_{bias,k}}(\bx)
=
\frac{1}{
(d_1-M_{L+1}+1) \cdot (d_2-M_{L+1}+1)
}
\cdot
	\sum_{ 	i \in \{1,\dots,d_1-M_{L+1}+1\}, \atop
	j \in \{1,\dots,d_2-M_{L+1}+1\} }
	 o_{(i,j),1,k}^{(L)} ,
\end{equation}
\begin{equation}
\label{se2eq3}
	o_{(i,j),s_2,k}^{(r)}
	=
	\sigma \left(
	\sum_{s_1=1}^{k_{r-1}}
	\sum_{\substack{t_1,t_2 \in \{1, \dots, M_r\}\\(i+t_1-1,j+t_2-1)\in D}}
	w_{t_1,t_2,s_1,s_2,k}^{(r)}
	\cdot
	o_{(i+t_1-1,j+t_2-1),s_1,k}^{(r-1)}
	+
	w_{s_2,k}^{(r)}
	\right)
\end{equation}
$((i,j)\in D=\{1,\dots,d_1\}\times\{1,\dots,d_2\}, s_2\in\{1,\dots,k_r\},
r \in \{1, \dots, L\})$
and
\begin{equation}
\label{se2eq4}
o_{(i,j),1,k }^{(0)} = x_{i,j}
\quad \mbox{for }
(i,j) \in D.
\end{equation}

Let $\bw$ be the vector of all the weights of the above network, i.e.,
$\bw$ contains $w_1, \dots, w_{K_n}$ together with all weights
$w_{t_1,t_2,s_1,s_2,k}^{(r)}$, $w_{s_2,k}^{(r)}$. We want to choose
$\bw$ such that the misclassification risk of $f_{\bw}$ is small.
To achieve this, we first estimate the a posteriori probability $\eta$
by a network $f_{\bw}$ which has a small empirical $L_2$ risk
\begin{equation}
\label{se2eq5}
\frac{1}{n} \sum_{i=1}^n |Y_i - f_{\bw}(\bX_i)|^2
\end{equation}
and then use the corresponding plug-in classificator for our image
classification problem.

Minimization of (\ref{se2eq5}) with respect to $\bw$
is a nonlinear minimization problem which, in general, cannot be
solved exactly. In the sequel we use gradient descent to obtain
an approximate solution to the minimization problem.

We start with a random initialization of $\bw$. We define $\bw^{(0)}$
by setting
\[
(\bw^{(0)})_k = 0 \quad (k=1, \dots, K_n)
\]
and by choosing all other components of $\bw^{(0)}$ as independent random variables sampled from
some uniform distributions. Here $w_{t_1,t_2,s_1,s_2,k}^{(r)}$ and $w_{s_2,k}^{(r)}$
are uniformly distributed on
$[-c_2 \cdot (\log n)^2, c_2 \cdot (\log n)^2]$
for $r=2, \dots, L$, and in case $r=1$ they are uniformly distributed
on
\[
[-c_3 \cdot (\log n)^2 \cdot n^\tau, c_3 \cdot (\log n)^2\cdot
n^\tau],
\]
where $\tau>0$ is a parameter of the estimate, which will be chosen
in Theorem \ref{th1} below.

Then we use gradient descent to define recursively weight vectors
$\bw^{(t)}$ for $t=1, \dots, t_n$. Here we add a regularization term
to the empirical $L_2$ risk (\ref{se2eq5}), i.e., we define
\begin{equation}
\label{se2eq6}
F_n(\bw)
=
\frac{1}{n} \sum_{i=1}^n |Y_i - f_{\bw}(\bX_i)|^2
+ c_4 \cdot \sum_{k=1}^{K_n} w_k^2,
\end{equation}
and apply gradient descent in order to minimize $F_n(\bw)$ with
respect to $\bw$, i.e., we compute
\begin{equation}
\label{se2eq7}
\bw^{(t)}
=
\bw^{(t-1)}
-
\lambda_n \cdot \nabla_{\bw} F_n(\bw^{(t-1)})
\quad
(t=1, \dots, t_n).
\end{equation}
Here $\lambda_n >0$ is the stepsize and $t_n \in \N$ is the number of
gradient
descent steps, and both will be chosen in Theorem \ref{th1} below.

Finally we define our image classifier $f_n$ as the plug-in
classifier corresponding to $f_{\bw^{(t_n)}}$, i.e., we set
\begin{equation}
\label{se2eq8}
f_n(\bx)=
\begin{cases}
	1, & \mbox{if } f_{\bw^{(t_n)}}(\bx) \geq \frac{1}{2}, \\
	0, & \mbox{elsewhere}.
\end{cases}
\end{equation}

\section{Main result}
\label{se3}
Our main result is the following bound  on the difference between
the misclassification risk of our estimator and the optimal
misclassification
risk.

\begin{theorem}
  \label{th1}
	Let $d_1, d_2, \kappa \in \N$ with $\kappa \leq \min\{d_1,d_2\}$.
	Let $(\bX,Y)$, $(\bX_1,Y_1)$, \dots, $(\bX_n,Y_n)$
	be independent and identically distributed
	$[0,1]^{\{1, \dots, d_1\} \times \{1, \dots, d_2\}} \times \{0,1\}$-valued
	random variables. Assume that the a posteriori probability
	$\eta(\bx)=\PROB\{Y=1|\bX=\bx\}$ satisfies
	a average-pooling model of size $\kappa^2$ with smoothness
        constraint $p \in \left[\frac{1}{2},1 \right]$.
	Choose
	\[
L \geq 2
\]
and $K_n \in \N$ such that
\begin{equation}
\label{th1eq1}
\frac{K_n}{n^{2 \cdot \kappa^2+7}}
\rightarrow \infty \quad (n \rightarrow \infty)
\end{equation}
and
\begin{equation}
\label{th1eq2}
\frac{K_n}{n^\rho} \rightarrow 0 \quad (n \rightarrow \infty)
\end{equation}
for some $\rho>0$
hold.
Choose $L_n \in \N$ with
\[
L_n \geq (\log n)^{6L+2} \cdot K_n^{3/2},
\]
set
\[
\lambda_n = \frac{1}{L_n}
\quad \mbox{and} \quad
t_n= \lceil c_5 \cdot (\log n) \cdot L_n \rceil,
\]
\[
\tau= \frac{1}{1 + \kappa^2}
\]
and
\begin{equation}
  \label{th1eq3}
M_1 =M_{L+1} = \kappa, \quad M_2= \dots = M_{L}=1, \quad
k_1= \dots = k_{L-1} = 2 \cdot \kappa^2
\quad \mbox{and} \quad
 k_0=k_L=1.
\end{equation}
Assume
\[
c_5 \geq \frac{1}{2 \cdot c_4}.
\]
Define the estimate as in Section \ref{se2}.
	Then we have for any $\epsilon>0$
	\begin{eqnarray*}
			&&\PROB\{f_n(\bX) \neq Y\}
			-
			\min_{f: [0,1]^{\{1, \dots, d_1\} \times \{1, \dots, d_2\}} \rightarrow \{0,1\}}
			\PROB\{f(\bX) \neq Y\}
\\
&& \leq
			c_6 \cdot n^{- \frac{1}{2 \cdot \kappa^2 +2}+\epsilon}
	\end{eqnarray*}
	for some constant $c_6 >0$ which does not depend on $d_1$,
        $d_2$  and $n$.\\

 \end{theorem}

\noindent
    {\bf Remark 1.} The above rate of convergence does not depend
    on the dimension $(d_1,d_2)$ of the image, instead it depends
    only on the parameter $\kappa^2$ (where $\kappa \leq \min\{d_1,d_2\}$)
    of the average--pooling model for $\eta$. Hence in case that the
    a posteriori probability $\eta$ satisfies an average--pooling model,
    our convolutional neural network estimate is able to circumvent
    the curse of dimensionality.

    \noindent
        {\bf Remark 2.} In the proof of Theorem \ref{th1} we show
        that a truncated version $\hat{\eta}_n$ of the convolutional
        neural network $f_{\bw^{(t_n)}}$ satisfies
        \[
        \EXP \int |\hat{\eta}_n(\bx)-\eta(\bx)|^2 \PROB_X(d \bx)
\leq
c_7 \cdot n^{- \frac{1}{\kappa^2 +1}+\epsilon}.
\]
According to Stone (1982), the optimal minimax rate of convergence
for estimation of a $d$-dimensional $(p,C)$--smooth regression function
is
\[
n^{-2p/(2p+d)}.
\]
Hence our truncated version $\hat{\eta}_n$ of the convolutional
neural network $f_{\bw^{(t_n)}}$  achieves a rate of convergence
which is close to the optimal minimax rate of convergence for estimation
of a $\kappa^2$-dimensional $(1/2,C)$--smooth regression function.

    \section{Proofs}
\label{se4}

\subsection{Auxiliary results}
\label{se4sub1}

\begin{lemma}
  \label{le1}
  Define $(\bX,Y)$, $(\bX_1,Y_1)$, \dots, $(\bX_n,Y_n)$, $\D_n$,
  $\eta$, and $f^*$ as in Subsection \ref{se1sub2}. Let
\[
\eta_n(\cdot)=\eta_n(\cdot,\D_n): [0,1]^{\{1, \dots, d_1\} \times \{1,
\dots, d_2\}} \rightarrow \R
\]
be an estimate of $\eta$ and set
\[
f_n(\bx)=
\begin{cases}
	1, & \mbox{if } \eta_n(\bx) \geq \frac{1}{2}, \\
	0, & \mbox{elsewhere}.
\end{cases}
\]
  Then
\begin{eqnarray*}
\PROB\{f_n(\bX) \neq Y|\D_n\}
-
\PROB\{ f^*(\bX) \neq Y\}
&\leq&
2 \cdot
\int |\eta_n(\bx)-\eta(\bx)| \, \PROB_{\bX}(d \bx)
\\
&\leq&
2 \cdot
\sqrt{
\int |\eta_n(x)-\eta(x)|^2  \PROB_{\bX} (dx)
  }
\end{eqnarray*}
  holds.
  \end{lemma}

\noindent
    {\bf Proof.}
    See Theorem 1.1 in Gy\"orfi et al. (2002).
    \hfill $\Box$

\begin{lemma}
  \label{le2}
  Let
  $F:\R^K \rightarrow \R_+$
  be a nonnegative differentiable function.
  Let
  $t \in \N$, $L>0$, $\ba_0 \in \R^K$ and set
  \[
  \lambda=
\frac{1}{L}
\]
and
\[
\ba_{k+1}=\ba_k - \lambda \cdot (\nabla_{\ba} F)(\ba_k)
\quad
(k \in \{0,1, \dots, t-1\}).
\]
Assume
\begin{equation}
  \label{le2eq1}
  \left\|
 (\nabla_{\ba} F)(\ba)
  \right\|
  \leq
  \sqrt{
2 \cdot t \cdot L \cdot \max\{ F(\ba_0),1 \}
    }
\end{equation}
for all $\ba \in \R^K$ with
$\| \ba - \ba_0\| \leq \sqrt{2 \cdot t \cdot \max\{ F(\ba_0),1 \} / L}$,
and
\begin{equation}
  \label{le2eq2}
\left\|
(\nabla_{\ba} F)(\ba)
-
(\nabla_{\ba} F)(\bb)
  \right\|
  \leq
  L \cdot \|\ba - \bb \|
\end{equation}
for all $\ba, \bb \in \R^K$ satisfying
\begin{equation}
  \label{le2eq3}
  \| \ba - \ba_0\| \leq \sqrt{8 \cdot \frac{t}{L} \cdot \max\{ F(\ba_0),1 \}}
  \quad \mbox{and} \quad
  \| \bb - \ba_0\| \leq \sqrt{8 \cdot \frac{t}{L} \cdot \max\{ F(\ba_0),1 \}}.
\end{equation}
Then we have
\[
\|\ba_k-\ba_0\| \leq
\sqrt{
2 \cdot \frac{k}{L} \cdot (F(\ba_0)-F(\ba_k))
}
\quad
 \mbox{for all }
 k \in \{1, \dots,t\},
\]
\[
\sum_{k=0}^{s-1}
\| \ba_{k+1}-\ba_k \|^2
\leq
\frac{2}{L}
 \cdot (F(\ba_0)-F(\ba_s))
\quad
 \mbox{for all }
 s \in \{1, \dots,t\}
 \]
 and
 \[
 F(\ba_k)
 \leq
 F(\ba_{k-1})
                -
                \frac{1}{2 L} \cdot
                \| \nabla_{\ba}  F(\ba_{k-1}) \|^2
                \quad
 \mbox{for all }
 k \in \{1, \dots,t\}.
\]
\end{lemma}

\noindent
    {\bf Proof.} The result follows from Lemma 2 in Braun et al. (2021)
    and its proof.
    \hfill $\Box$

\begin{lemma}\label{le3}
	Let $\sigma: \R \rightarrow \R$ be bounded and differentiable, and assume that
	its derivative is bounded.
	Let
	$t_n \geq L_n$,
	$\gamma_n^* \geq 1$, $B_n \geq 1$,
	\begin{equation}
		\label{le3eq1}
		|w_k| \leq \gamma_n^* \quad (k=1, \dots, K_n),
	\end{equation}
	\begin{equation}
		\label{le3eq2}
		|w_{s_2,k}^{(r)}| \leq B_n
		\text{ and }
		|w_{t_1,t_2,s_1,s_2,k}^{(r)}| \leq B_n
		\quad
		\mbox{for } r=2, \dots, L
	\end{equation}
	and
	\begin{equation}
		\label{le3eq3}
		\|\bw-\bv\|_\infty^2 \leq \frac{2t_n}{L_n} \cdot \max\{ F_n(\bv),1 \}.
	\end{equation}
	Assume $X_1, \dots, X_n \in [0,1]^{\{1,\dots,d_1\}\times\{1,\dots,d_2\}}$ and define
	$F_n$ by (\ref{se2eq6}), where parameters $L\in\N$, $M_1,\dots,M_{L+1}\in\N$
        and $k_0, \dots, k_L \in \N$
        of the convolutional neural network used in \eqref{se2eq6} satisfy
        $L\geq 2$, $M_2=\dots=M_{L}=1$, $M_1=M_{L+1}=\kappa$,
        $k_1=\dots=k_{L-1}=2 \cdot \kappa^2$ and $k_0=k_L=1$.
	
	Then we have
	\[
	\| (\nabla_\bw F_n)(\bw) \|
	\leq
	c_{8} \cdot K_n^{3/2} \cdot B_n^{2L} \cdot (\gamma_n^*)^2 \cdot \sqrt{\frac{t_n}{L_n} \cdot \max\{F_n(\bv),1\}}.
	\]
\end{lemma}
\noindent
    {\bf Proof.} Because of $M_2=\dots=M_{L}=1$, $M_1=M_{L+1}=\kappa$,
    $k_1=\dots=k_{L-1}=2 \cdot \kappa^2$ and $k_0=1$
    we have
    \[
    f_{\bw}(\bx) = \sum_{k=1}^{K_n} w_k \cdot f_{\bw_k, \bw_{bias,k}}(\bx)
    \]
where for $k \in \{1, \dots, K_n\}$
\[
f_{\bw_k, \bw_{bias,k}}(\bx)
=
\frac{1}{
(d_1-\kappa+1) \cdot (d_2-\kappa+1)
}
\cdot
	\sum_{ 	i \in \{1,\dots,d_1-\kappa+1\}, \atop
	j \in \{1,\dots,d_2-\kappa+1\} }
	 o_{(i,j),1,k}^{(L)} ,
\]
\[
	o_{(i,j),s_2,k}^{(r)}
	=
	\sigma \left(
	\sum_{s_1=1}^{2 \cdot \kappa^2}
	w_{1,1,s_1,s_2,k}^{(r)}
	\cdot
	o_{(i,j),s_1,k}^{(r-1)}
	+
	w_{s_2,k}^{(r)}
	\right)
\]
$((i,j) \in D, s_2 \in \{1, \dots, k_r\}, r \in \{2, \dots, L\})$
and
\[
	o_{(i,j),s_2,k}^{(1)}
	=
	\sigma \left(
	\sum_{\substack{t_1,t_2 \in \{1, \dots, \kappa\}\\(i+t_1-1,j+t_2-1)\in D}}
	w_{t_1,t_2,1,s_2,k}^{(1)}
	\cdot
	x_{i+t_1-1,j+t_2-1}
	+
	w_{s_2,k}^{(1)}
	\right)
\]
for
$(i,j) \in D$ and $s_2 \in \{1, \dots, 2 \cdot \kappa^2\}$.

Following to the proof of Lemma 2 in Drews and Kohler (2022), we get
\begin{eqnarray*}
  &&
  \|(\nabla_{\bw} F_n)(\bw)\|^2
  \\
  &&
  =
  \sum_{k=1}^{K_n} \left(
  \frac{2}{n} \sum_{i=1}^n (f_{\bw}(X_i)-Y_i) \cdot \frac{\partial f_{\bw}}{
  \partial w_k}(X_i) + c_4 \cdot 2 \cdot w_k
  \right)^2
  \\
  &&
  \quad
  +\sum_{k=1}^{K_n}
  \sum_{r=1}^L
  \sum_{s_1 \in \{1, \dots, k_{r-1}\}, s_2 \in \{1, \dots, k_r\}
    \atop
    t_1, t_2 \in \{1, \dots, M_r\}}
  \left(
  \frac{2}{n} \sum_{i=1}^n (f_{\bw}(X_i)-Y_i) \cdot \frac{\partial f_{\bw}}{
    \partial w_{t_1,t_2,s_1,s_2,k}^{(r)}}(X_i)
  \right)^2
  \\
  &&
  \quad
  +
  \sum_{k=1}^{K_n}
  \sum_{r=1}^L
  \sum_{s_2 \in \{1, \dots, k_{r}\}
   }
  \left(
  \frac{2}{n} \sum_{i=1}^n (f_{\bw}(X_i)-Y_i) \cdot \frac{\partial f_{\bw}}{
    \partial w_{s_2,k}^{(r)}}(X_i)
  \right)^2
  \\
  &&
  \leq
  c_8\cdot \kappa^4 \cdot K_n\cdot L\cdot
  \max \Bigg(
  \max_{k,i} 	\left(
	\frac{\partial f_{\bw}}{\partial w_{k}(\bX_i)}
	\right)^2,
  \max_{t_1,t_2,s_1,s_2,k,r,i}
	\left(
	\frac{\partial f_{\bw}}{\partial w_{t_1,t_2,s_1,s_2,k}^{(r)}}(\bX_i)
	\right)^2,
        \\
        &&
        \quad
 \max_{s_2,k,r,i}
	\left(
	\frac{\partial f_{\bw}}{\partial w_{s_2,k}^{(r)}}(\bX_i)
	\right)^2
        \Bigg)
	\cdot \frac{1}{n} \cdot \sum_{i=1}^{n}(f_{\bw}(\bX_i)-Y_i)^2\\
        && \quad
	+8\cdot c_4^2\cdot K_n\cdot(\gamma^*_n)^2.
  \end{eqnarray*}
Next we calculate the derivatives
\[
\frac{\partial f_{\bw}}{\partial w_{k}}(\bx), \quad
\frac{\partial f_{\bw}}{\partial w_{t_1,t_2,s_1,s_2,k}^{(r)}}(\bx)
\quad \mbox{and} \quad
\frac{\partial f_{\bw}}{\partial w_{s_2,k}^{(r)}}(\bx).
\]
We have
\[
\frac{\partial f_{\bw}}{\partial w_{k}}(\bx) = f_{\bw_k, \bw_{bias,k}}(\bx).
\]
Furthermore
\[
\frac{\partial f_{\bw}}{\partial w_{t_1,t_2,s_1,s_2,k}^{(r)}}(\bx)
=
w_k \cdot
\frac{1}{
(d_1-\kappa+1) \cdot (d_2-\kappa+1)
}
\cdot
	\sum_{ 	i \in \{1,\dots,d_1-\kappa+1\}, \atop
	j \in \{1,\dots,d_2-\kappa+1\} }
\frac{\partial o_{(i,j),1,k}^{(L)}}{\partial w_{t_1,t_2,s_1,s_2,k}^{(r)}}
\]
and
\[
\frac{\partial f_{\bw}}{\partial w_{s_2,k}^{(r)}}(\bx)
=
w_k \cdot
\frac{1}{
(d_1-\kappa+1) \cdot (d_2-\kappa+1)
}
\cdot
	\sum_{ 	i \in \{1,\dots,d_1-\kappa+1\}, \atop
	j \in \{1,\dots,d_2-\kappa+1\} }
\frac{\partial o_{(i,j),1,k}^{(L)}}{\partial w_{s_2,k}^{(r)}}.
\]
In the following we calculate the derivatives
\[
\frac{\partial o^{(L)}_{(i,j),1,k}}{\partial w_{t_1,t_2,s_1,s_2,k}^{(r)}}.
\]
In the $L$--th layer we have
\begin{align*}
	&\frac{\partial o^{(L)}_{(i,j),1,k}}{\partial w_{1,1,s_1,1,k}^{(L)}}
	\\&=
	o^{(L-1)}_{(i,j),s_1,k}\cdot
	\sigma^{\prime}\Bigg(
        	\sum_{s=1}^{2 \cdot \kappa^2}
	w_{1,1,s,1,k}^{(L)}
	\cdot
	o_{(i,j),s,k}^{(L-1)}
	+
	w_{1,k}^{(L)}
	\Bigg)
\end{align*}
for $(i,j) \in D$, $s_1 \in \{1, \dots, 2 \cdot \kappa^2\}$ and $k\in\{1,\dots,K_n\}$.
For $k\in\{1,\dots,K_n\}$, $r\in\{1,\dots,L-1\}$,
$s_1 \in \{1, \dots, M_r\}$, $s_2 \in \{1, \dots, M_{r+1}\}$ and
$t_1, t_2 \in \{1, \dots, M_r\}$ we get by using
the chain rule
\begin{eqnarray*}
  &&
  \frac{\partial o^{(L)}_{(i,j),1,k}}{\partial w_{t_1,t_2,s_1,s_2,k}^{(r)}} \\
  &&
  =
  	\sigma^\prime \left(
	\sum_{s=1}^{2 \cdot \kappa^2}
	w_{1,1,s,1,k}^{(L)}
	\cdot
	o_{(i,j),s,k}^{(L-1)}
	+
	w_{1,k}^{(L)}
	\right)
        \cdot
	\sum_{s^{(L)}=1}^{2 \cdot \kappa^2}
	w_{1,1,s^{(L)},1,k}^{(L)}
	\cdot
          \frac{\partial o_{(i,j),s^{(L)},k}^{(L-1)}  }{\partial w_{t_1,t_2,s_1,s_2,k}^{(r)}}
	  \\
          &&
        =
        \dots
        \\
        &&
        =
        	\sum_{s^{(L)}=1}^{2 \cdot \kappa^2}
	        \sum_{s^{(L-1)}=1}^{2 \cdot \kappa^2}
                \dots
                	\sum_{s^{(r+2)}=1}^{2 \cdot \kappa^2}
  	\sigma^\prime \left(
	\sum_{s=1}^{2 \cdot \kappa^2}
	w_{1,1,s,1,k}^{(L)}
	\cdot
	o_{(i,j),s,k}^{(L-1)}
	+
	w_{1,k}^{(L)}
	\right)
        \cdot
	w_{1,1,s^{(L)},1,k}^{(L)} \cdot
        \\
        &&
        \quad
  	\sigma^\prime \left(
\sum_{s=1}^{2 \cdot \kappa^2}
	w_{1,1,s,s^{(L)},k}^{(L-1)}
	\cdot
	o_{(i,j),s,k}^{(L-2)}
	+
	w_{s^{(L)},k}^{(L-1)}
	\right)
        \cdot	w_{1,1,s^{(L-1)},s^{(L)},k}^{(L-1)} \cdots
        \\
        &&
        \quad
        \sigma^\prime \left(
\sum_{s=1}^{2 \cdot \kappa^2}
	w_{1,1,s,s^{(r+2)},k}^{(r+1)}
	\cdot
	o_{(i,j),s,k}^{(r)}
	+
	w_{s^{(r+2)},k}^{(r+1)}
	\right)
        \cdot
        w_{1,1,s_2,s^{(r+2)},k}^{(r+1)} \cdot
        \\
        &&
        \quad
        \Bigg( I_{\{r>1\}} \cdot
        \sigma^\prime \left(
\sum_{s=1}^{2 \cdot \kappa^2}
	w_{1,1,s,s_2,k}^{(r)}
	\cdot
	o_{(i,j),s,k}^{(r-1)}
	+
	w_{s_2,k}^{(r)}
	\right)
        \cdot
        o_{(i,j),s_1,k}^{(r-1)}
        \\
        &&
        \quad
        +
        I_{\{r=1\}} \cdot
        	\sigma^\prime \left(
	\sum_{\substack{\tilde{t}_1,\tilde{t}_2 \in \{1, \dots, \kappa\}\\(i+\tilde{t}_1-1,j+\tilde{t}_2-1)\in D}}
	w_{\tilde{t}_1,\tilde{t}_2,1,s_2,k}^{(1)}
	\cdot
	x_{i+\tilde{t}_1-1,j+\tilde{t}_2-1}
	+
	w_{s_2,k}^{(1)}
	\right)
	\cdot
	x_{i+t_1-1,j+t_2-1}
\Bigg),
\end{eqnarray*}
where we have set $x_{i,j}=0$ for $(i,j) \notin D$.
For the partial derivatives with respect to $w_{s_2,k}^{(r)}$ we
can easily show a similar result.

Using the assumptions of Lemma \ref{le3} we can conclude
\begin{eqnarray*}
  &&
    \max \Bigg(
  \max_{k,i} 	\left(
	\frac{\partial f_{\bw}}{\partial w_{k}(\bX_i)}
	\right)^2,
  \max_{t_1,t_2,s_1,s_2,k,r,i}
	\left(
	\frac{\partial f_{\bw}}{\partial w_{t_1,t_2,s_1,s_2,k}^{(r)}}(\bX_i)
	\right)^2,
 \max_{s_2,k,r,i}
	\left(
	\frac{\partial f_{\bw}}{\partial w_{s_2,k}^{(r)}}(\bX_i)
	\right)^2
        \Bigg)
\\
&&
\leq
c_9 \cdot \kappa^{4L} \cdot
\max\{ \|\sigma^\prime\|_\infty^{2L},1\} \cdot
\max\{ \|\sigma\|_\infty^{2},1\} \cdot B_n^{2L} \cdot (\gamma_n^*)^2.
\end{eqnarray*}

Next we show that for any $\bx \in [0,1]^{\{1,\dots,d_1\}\times\{1,\dots,d_2\}}$
\[
|f_\bw(x)-f_\bv(x)| \leq
K_n \cdot \max\{\|\sigma'\|_\infty^L,1\} \cdot \gamma_n^* \cdot (4\cdot
\kappa^4 +1)^L \cdot B_n^L\cdot \max\{\|\sigma\|_\infty,1\} \cdot \|\bw-\bv\|_\infty.
\]
This follows from
\begin{align*}
	&|f_{\bw}(\bx)-f_{\bv}(\bx)|\\
	&=\left|\sum_{k=1}^{K_n}w_k\cdot f_{\bw_{k},\bw_{bias,k}}(\bx)-\sum_{k=1}^{K_n}v_k\cdot f_{\bv_{k},\bv_{bias,k}}(\bx)\right|\\
	&\leq K_n\cdot\max_{k\in\{1,\dots,K_n\}}
	\left\{
	|w_k-v_k|\cdot \|\sigma\|_{\infty},
	\gamma_{n}^{*}\cdot|f_{\bw_{k},\bw_{bias,k}}(\bx)-f_{\bv_{k},\bv_{bias,k}}(\bx)|
	\right\}\\
	&
	\leq K_n\cdot\max_{k\in\{1,\dots,K_n\}}
	\left\{
	|w_k-v_k|\cdot \|\sigma\|_{\infty},
	\gamma_{n}^{*}\cdot
	\max_{i\in\{1,\dots,d_1-\kappa+1\}\atop j\in\{1,\dots,d_2-\kappa+1\}}\left|o_{(i,j),1,k}^{(L)}-\bar{o}_{(i,j),1,k}^{(L)}\right|
	\right\}
\end{align*}
(where $\bar{o}_{(i,j),s_2,k}^{(r)}$ is defined by replacing
in the definition of $o_{(i,j),s_2,k}^{(r)}$ $(\bw_k,\bw_{bias,k})$
by $(\bv_k,\bv_{bias,k})$)
and that for $r\in\{1,\dots,L\}$ we have
\begin{align*}
	&\left|o_{(i,j),s_2,k}^{(r)}-\bar{o}_{(i,j),s_2,k}^{(r)}\right|\\
	&\leq
	\max\{\|\sigma^{\prime}\|_{\infty}^r,1\}
	\cdot (4\cdot \kappa^4+1)^r\cdot B_n^r
	\cdot\max\{\|\sigma\|_{\infty},1\}\\
	&\quad
        \cdot
        \max\left\{
	\max_{\tilde{r} \in\{1,\dots,L\}\atop \tilde{t}_1,\tilde{t}_2,\tilde{s}_1,\tilde{s}_2, \tilde{k}}
	\left|w^{(\tilde{r})}_{\tilde{t}_1,\tilde{t}_2,\tilde{s}_1,\tilde{s}_2, \tilde{k}}
        -v^{(\tilde{r})}_{\tilde{t}_1,\tilde{t}_2,\tilde{s}_1,\tilde{s}_2, \tilde{k}}\right|,
        \max_{\tilde{r} \in\{1,\dots,L\}\atop \tilde{s}_2, \tilde{k}}
        \left|
        w_{\tilde{s}_2, \tilde{k}}^{(\tilde{r})}
        -
        v_{\tilde{s}_2, \tilde{k}}^{(\tilde{r})}
        \right|
        \right\},
\end{align*}
which we can easily be shown by induction on $r$
(cf., e.g., proof of Lemma 5 in
Kohler and Krzy\.zak (2021) for a related proof).

This implies
\begin{eqnarray*}
	&&
	\frac{1}{n}
	\sum_{s=1}^n
	(Y_s - f_\bw (X_s))^2
	\\
	&&
	\leq
	2 \cdot F_n(\bv) +       \frac{2}{n}
	\sum_{s=1}^n
	(f_\bv(X_s) - f_\bw (X_s))^2
	\\
	&&
	\leq 2 \cdot F_n(\bv) +
	2 \cdot
	K_n^2 \cdot \max\{\|\sigma'\|_\infty^{2L},1\} \cdot (\gamma_n^*)^2 \cdot
        (4 \cdot \kappa^4+1)^{2L} \cdot B_n^{2L}
	\\
	&&
	\quad
	\cdot \max\{\|\sigma\|_\infty,1\}^2
	\cdot \frac{2t_n}{L_n} \cdot \max\{F_n(\bv),1\}.
\end{eqnarray*}
The proof is completed by putting together the above results.
\hfill $\Box$

\begin{lemma}
	\label{le4}
	Let $\sigma: \R \rightarrow \R$ be bounded and differentiable, and assume that its derivative
	is
	Lipschitz continuous and bounded.
	Let
	$t_n \geq L_n$,
	$\gamma_n^* \geq 1$, $B_n \geq 1$ and assume
	\begin{equation}
		\label{le4eq1}
		|\max\{ (\bw_1)_k, (\bw_2)_k \}| \leq \gamma_n^* \quad (k=1,
		\dots, K_n),
	\end{equation}
	
	\begin{equation}
		\label{le4eq2}
		|\max\{(\bw_1)_{s_1,s_2,t_1,t_2,k}^{(r)},(\bw_2)_{s_1,s_2,t_1,t_2,k}^{(r)}\}| \leq B_n
		\quad
		\mbox{for } r=2, \dots, L
	\end{equation}
	and
	\begin{equation}
		\label{le4eq3}
		\|\bw_2-\bv\|^2 \leq 8 \cdot \frac{t_n}{L_n} \cdot \max\{ F_n(\bv),1 \}.
	\end{equation}
	Assume $X_1, \dots, X_n \in [0,1]^{\{1,\dots,d_1\}\times\{1,\dots,d_2\}}$ and define
	$F_n$ by (\ref{se2eq6}), where the parameters $L\in\N$,
        $M_1,\dots,M_{L+1}\in\N$ and $k_0, \dots, k_L \in \N$
        of the convolutional neural network used in \eqref{se2eq6} satisfy $L\geq 2$, $M_2=\dots=M_{L}=1$, $M_1=M_{L+1}=\kappa$,
        $k_1=\dots=k_{L-1}=2 \cdot \kappa^2$ and $k_0=k_L=1$.
	
	Then we have
	\begin{eqnarray*}
		&&
		\| (\nabla_\bw F_n)(\bw_1) - (\nabla_\bw F_n)(\bw_2) \| \\
		&&
		\leq
		c_{10} \cdot \max \{\sqrt{F_n(\bv)},1\} \cdot (\gamma_n^*)^{2} \cdot B_n^{3L-1} \cdot  K_n^{3/2} \cdot \sqrt{\frac{t_n}{L_n}} \cdot \|\bw_1-\bw_2\|.
	\end{eqnarray*}
\end{lemma}

\noindent
{\bf Proof.}
    Using the formulas for the partial derivatives derived in the proof
    of Lemma \ref{le3} the assertion follows as in the proof of
    Lemma 3 in Drews and Kohler (2022).
\hfill $\Box$

\begin{lemma}
  \label{le5a}
  Let $\alpha \geq 1$, $\beta>0$ and let $A,B,C \geq 1$.
  Let $\sigma:\R \rightarrow \R$ be $k$-times differentiable
  such that all derivatives up to order $k$ are bounded on $\R$.
Let $d_1, d_2, \kappa \in \N$ such that $\kappa \leq
\min\{d_1,d_2\}$.
  Let $\F$
  be the set of all functions $f_{\bw}$
defined by (\ref{se2eq1})--(\ref{se2eq4}) with
\begin{equation}
\label{le5aeq1}
M_1 =M_{L+1}= \kappa, \quad M_2= \dots = M_{L}=1, \quad
k_1= \dots = k_{L-1} = 2 \cdot \kappa^2
\quad \mbox{and} \quad
 k_0=k_L=1,
\end{equation}
where the weights
satisfy
  \begin{equation}
    \label{le5aeq2}
    \sum_{j=1}^{K_n} |w_k| \leq C,
    \end{equation}
  \begin{equation}
    \label{le5aeq3}
\max\{    |w_{t_1,t_2,s_1,s_2,k}^{(l)}|,
|w_{s_2,k}^{(l)}|
\} \leq B \quad (k \in \{1, \dots, K_n\},
     l \in \{2, \dots, L\})
    \end{equation}
and
  \begin{equation}
    \label{le5aeq4}
\max\{    |w_{t_1,t_2,s_1,s_2,k}^{(1)}|,
|w_{s_2,k}^{(1)}|
\} \leq A \quad (k \in \{1, \dots, K_n\}).
  \end{equation}
  Then we have for any $1 \leq p < \infty$, $0 < \epsilon < \beta$ and
  $\bx_1^n \in [0,1]^{\{1, \dots, d_1\} \times \{1, \dots, d_2\}}$
  \begin{eqnarray*}
    &&
  \Nu_p \left(
\epsilon, \{ T_\beta f  \, : \, f \in \F \}, \bx_1^n
\right)
\\
&&
\leq
\left(
c_{11} \cdot \frac{\beta^p }{\epsilon^p}
\right)^{
c_{12}  \cdot A^{\kappa^2} \cdot B^{(L-1) \cdot \kappa^2}
   \left(\frac{ C}{\epsilon}\right)^{\kappa^2/k} + c_{13}
  }.\\
  \end{eqnarray*}

  \end{lemma}
For the proof of Lemma \ref{le5a} we need the following result from
Kohler and Krzy\.zak (2022).

\begin{lemma}
  \label{le5}
  Let $\alpha \geq 1$, $\beta>0$ and let $A,B,C \geq 1$.
  Let $\sigma:\R \rightarrow \R$ be $k$-times differentiable
  such that all derivatives up to order $k$ are bounded on $\R$.
Let $d \in \N$ and let $1 \leq d^* \leq d$. For $x=(x^{(1)}, \dots,
x^{(d)})$ and $I \subset \{1, \dots, d\}$ set $x_I = (x^{(i)})_{i \in I}$.
  Let $\F$
  be the set of all functions
\[
f_{\bw}(x) = \sum_{I \subset \{1, \dots, d\}, |I| = d^*} f_{\bw_I}(x_I)
\]
where the
$f_{\bw_I}(z)$ are defined for $z \in \R^{d^*}$ by
\begin{equation}\label{le5eq1}
f_{\bw_I}(z) = \sum_{k=1}^{K_n} {(\bw_I)}_{1,1,k}^{(L)} \cdot f_{{\bw_I},k,1}^{(L)}(z)
\end{equation}
for some ${(\bw_I)}_{1,1,1}^{(L)}, \dots, {(\bw_I)}_{1,1,K_n}^{(L)} \in \mathbb{R}$, where
$f_{{\bw_I},j,1}^{(L)}$ are recursively defined by
\begin{equation}
  \label{le5eq2}
f_{{\bw_I},k,i}^{(l)}(z) = \sigma\left(\sum_{j=1}^{r} {(\bw_I)}_{k,i,j}^{(l-1)}\cdot f_{{\bw_I},k,j}^{(l-1)}(z) + {(\bw_I)}_{k,i,0}^{(l-1)} \right)
\end{equation}
for some ${(\bw_I)}_{k,i,0}^{(l-1)}, \dots, {(\bw_I)}_{k,i, r}^{(l-1)} \in \mathbb{R}$
$(l=2, \dots, L)$
and
\begin{equation}
  \label{le5eq3}
f_{{\bw_I},k,i}^{(1)}(z) = \sigma \left(\sum_{j=1}^{d^*} {(\bw_I)}_{k,i,j}^{(0)}\cdot z^{(j)} + {(\bw_I)}_{k,i,0}^{(0)} \right)
\end{equation}
for some ${(\bw_I)}_{k,i,0}^{(0)}, \dots, {(\bw_I)}_{k,i,d}^{(0)} \in \mathbb{R}$, and
where ${\bw_I}$ denotes the vector of all weights ${(\bw_I)}_{1,1,j}^{(L)}$ and
${(\bw_I)}_{k,i,j}^{(l)}$ $(l=1, \dots, L-1)$, and
where for each $I \subseteq \{1, \dots, d\}$, $|I|=d^*$
the weight vector ${\bw_I}$
  satisfies
  \begin{equation}
    \label{le5eq4}
    \sum_{j=1}^{K_n} |{(\bw_I)}_{1,1,j}^{(L)}| \leq C,
    \end{equation}
  \begin{equation}
    \label{le5eq5}
    |{(\bw_I)}_{k,i,j}^{(l)}| \leq B \quad (k \in \{1, \dots, K_n\},
    i,j \in \{1, \dots, r\}, l \in \{1, \dots, L-1\})
    \end{equation}
and
  \begin{equation}
    \label{le5eq6}
    |{(\bw_I)}_{k,i,j}^{(0)}| \leq A \quad (k \in \{1, \dots, K_n\},
    i \in \{1, \dots, r\}, j \in \{1, \dots,d^*\}).
  \end{equation}
  Then we have for any $1 \leq p < \infty$, $0 < \epsilon < \beta$ and
  $x_1^n \in [-\alpha,\alpha]^d$
  \begin{eqnarray*}
    &&
  \Nu_p \left(
\epsilon, \{ T_\beta f  \, : \, f \in \F \}, x_1^n
\right)
\\
&&
\leq
\left(
c_{14} \cdot \frac{\beta^p }{\epsilon^p}
\right)^{
c_{15} \cdot \alpha^{d^*} \cdot A^{d^*} \cdot B^{(L-1) \cdot d^*} \left(\frac{C}{\epsilon}\right)^{d^*/k} + c_{16}
  }.\\
  \end{eqnarray*}

  \end{lemma}

\noindent
{\bf Proof.} See Lemma 8 in Kohler and Krzy\.zak (2022) and its proof.
\hfill $\Box$

\noindent
    {\bf Proof of Lemma \ref{le5a}.} Set
    \[
I = \{0, \dots, \kappa-1\} \times \{0, \dots, \kappa-1\}.
    \]
 Using (\ref{le5aeq1}) it is easy
to
see that we can find weight vectors $\bw_{k,(i,j)}$
such that
\[
f_{\bw_k,\bw_{bias,k}}(\bx)
=
\frac{1}{(d_1-\kappa+1) \cdot (d_2 - \kappa +1)}
\cdot
\sum_{i \in \{1, \dots, d_1-\kappa+1\}, \atop
j \in \{1, \dots, d_2-\kappa+1\}}
f_{\bw_{k,(i,j)},k,1}^{(L)}(\bx_{(i,j)+I})
\]
holds. This implies
\[
f_{\bw}(\bx)
=
\sum_{i \in \{1, \dots, d_1-\kappa+1\}, \atop
j \in \{1, \dots, d_2-\kappa+1\}}
\sum_{k=1}^{K_n}
\frac{1}{(d_1-\kappa+1) \cdot (d_2 - \kappa +1)}
w_k
\cdot
f_{\bw_{k,(i,j)},k,1}^{(L)}(\bx_{(i,j)+I}).
\]
Furthermore from (\ref{le5aeq2}), (\ref{le5aeq3}) and (\ref{le5aeq4}) we can conclude
that (\ref{le5eq4}),  (\ref{le5eq5}) and (\ref{le5eq6}) hold.
Application of Lemma \ref{le5} with $d=d_1 \cdot d_2$ and
$d^*=\kappa^2$
yields the desired result.
\hfill $\Box$

\begin{lemma}
\label{le6a}
Let $d_1,d_2,\kappa \in \N$ with $\min\{d_1,d_2\} \geq \kappa$
and set
\[
I= \{0,\dots,\kappa-1\} \times \{0,\dots,\kappa-1\}.
\]
Let $1/2 \leq p \leq 1$, $C>0$,
let $f:[0,1]^{(1,1)+I} \rightarrow \R$ be a $(p,C)$--smooth function,
let $X$ be $[0,1]^{\{1, \dots, d_1\} \times \{1, \dots, d_2 \}}$-valued
random vector and
for
$x \in [0,1]^{\{1, \dots, d_1\} \times \{1, \dots, d_2 \}}$
set
\[
	m(\bx)=
\frac{1}{(d_1 -\kappa+1) \cdot (d_2 - \kappa +1)} \cdot
		\sum_{
			(i,j) \in \Z^2 \, : \,
			(i,j)+I \subseteq \{1, \dots, d_1\} \times \{1, \dots, d_2\}
		}
		f\left(
		\bx_{(i,j)+I}
		\right).
\]
Let $l \in \N$, $0<\delta<1/2$ with
\begin{equation}
\label{le6aeq2}
c_{17} \cdot \delta \leq \frac{1}{2^l} \leq c_{18} \cdot \delta
\end{equation}
 and let $L,s \in \N$ with $L \geq 2$,
set
\[
M_1 = M_{L+1}=\kappa, \quad M_2= \dots = M_{L}=1, \quad
k_1= \dots = k_{L-1} = 2 \cdot \kappa^2
\quad \mbox{and} \quad
 k_0=k_L=1,
\]
and let
\[
\tilde{K}_n \geq \left( l \cdot (2^l+1)^{2 \kappa^2}+1 \right)^3.
\]
Let
\[
f_{\bw}(\bx)=\sum_{k=1}^{\tilde{K}_n} w_k \cdot f_{\bw_k,\bw_{bias,k}}(\bx)
\]
where $f_{\bw_k,\bw_{bias,k}}(\bx)$ is defined by (\ref{se2eq2}), (\ref{se2eq3}) and
(\ref{se2eq4}).
Then there exist
\[
w_k, w_{t_1,t_2,s_1,s_2,k}^{(l)}, w_{s_2,k}^{(l)}\in
[-c_2 \cdot (\log n)^2, c_2 \cdot (\log n)^2]
\quad (l=2, \dots, L, k=1, \dots, \tilde{K}_n)
\]
and
\[
w_{t_1,t_2,s_1,s_2,k}^{(1)}, w_{s_2,k}^{(1)}
\in \left[-\frac{8 \cdot \kappa^2 \cdot (\log n)^2}{\delta}, \frac{8 \cdot \kappa^2 \cdot
      (\log n)^2}{\delta} \right]
\quad (k=1, \dots, \tilde{K}_n)
\]
such that
for all $\bar{\bw}$ satisfying
$|\bar{w}_{t_1,t_2,s_1,s_2,k}^{(l)}-w_{t_1,t_2,s_1,s_2,k}^{(l)}| \leq
\log n$
and
$|\bar{w}_{s_2,k}^{(l)}-w_{s_2,k}^{(l)}| \leq
\log n$
$(l=1, \dots, L)$ we have for $n$ sufficiently large
\begin{eqnarray}
\label{le6aeq3}
&&
\int
|
\sum_{k=1}^{\tilde{K}_n}
w_{k} \cdot f_{\bar{\bw}_k,\bar{\bw}_{bias,k}}^{(L)}(\bx)-m(x)|^2 \PROB_{\bX} (d\bx)
\nonumber \\
&&
\leq
c_{19} \cdot \left(
l^2 \cdot
(d_1-\kappa+1) \cdot (d_2-\kappa+1)
 \cdot \delta + \delta^{2p}
+
\frac{l \cdot (2^l+1)^{2 \cdot \kappa^2}}{n^s}
\right)
,
\end{eqnarray}
\begin{equation}
\label{le6aeq1}
|
\sum_{k=1}^{\tilde{K}_n}
w_k \cdot f_{\bar{\bw}_k,\bar{\bw}_{bias,k}}^{(L)}(\bx) |\leq
c_{20}
\cdot
\left( 1 +
\frac{(2^l+1)^{2 \cdot \kappa^2}}{n^s}
\right)
\quad (\bx \in [0,1]^{\{1, \dots, d_1\} \times \{1, \dots, d_2 \}})
\end{equation}
and
\begin{equation}
\label{le6aeq4}
\sum_{k=1}^{\tilde{K}_n} |w_k|^2 \leq \frac{c_{21}}{2^{2    \cdot \kappa^2
    \cdot l}} .
\end{equation}
\end{lemma}
To prove Lemma \ref{le6a} we need the following result from
Kohler and Krzy\.zak (2022).

\begin{lemma}
\label{le6}
Let $1/2 \leq p \leq 1$, $C>0$,
let $f:\Rd \rightarrow \R$ be a $(p,C)$--smooth function,
let $N \in \N$ and let $Z_1$, \dots, $Z_N$ be $[0,1]^d$-valued
random vectors.
Let $l \in \N$, $0<\delta<1/2$ with
\begin{equation}
\label{le6eq2}
c_{22} \cdot \delta \leq \frac{1}{2^l} \leq c_{23} \cdot \delta
\end{equation}
 and let $L,r,s \in \N$ with
\[
L \geq 2 \quad \mbox{and} \quad r \geq 2d
\]
and let
\[
\tilde{K}_n \geq \left( l \cdot (2^l+1)^{2d}+1 \right)^3
\]
Define $f_{\bw,k,1}^{(L)}$ by (\ref{le5eq2}) and (\ref{le5eq3}) with
$d^*$ replaced by $d$.
Then there exist
\[
{\bw}_{k,i,j}^{(l)} \in
[-c_2 \cdot (\log n)^2, c_2 \cdot (\log n)^2]
\quad (l=1, \dots, L, k=1, \dots \tilde{K}_n)
\]
and
\[
{\bw}_{k,i,j}^{(0)}
\in \left[-\frac{8 \cdot d \cdot (\log n)^2}{\delta}, \frac{8 \cdot d \cdot
      (\log n)^2}{\delta} \right]
\quad (k=1, \dots, \tilde{K}_n).
\]
such that
for all $\bar{\bw}$ satisfying
$|\bar{w}_{i,j,k}^{(l)}-\bw_{i,j,k}^{(l)}| \leq \log n$
$(l=0, \dots, L-1)$ we have for $n$ sufficiently large
\begin{eqnarray}
\label{le6eq3}
&&
\max_{i=1, \dots, N}
\int
|
\sum_{k=1}^{\tilde{K}_n}
w_{1,1,k}^{(L)} \cdot f_{\bar{\bw},k,1}^{(L)}(x)-f(x)|^2 \PROB_{Z_i} (dx)
\nonumber \\
&&
\hspace*{2cm}
\leq
c_{24} \cdot \left(
l^2 \cdot N \cdot \delta + \delta^{2p}
+
\frac{l \cdot (2^l+1)^{2d}}{n^s}
\right)
,
\end{eqnarray}
\begin{equation}
\label{le6eq1}
|
\sum_{k=1}^{\tilde{K}_n}
w_{1,1,k}^{(L)} \cdot f_{\bar{\bw},k,1}^{(L)}(x)| \leq
c_{25}
\cdot
\left( 1 +
\frac{(2^l+1)^{2d}}{n^s}
\right)
\quad (x \in [0,1]^d)
\end{equation}
and
\begin{equation}
\label{le6eq4}
\sum_{k=1}^{\tilde{K}_n} |w_{1,1,k}^{(L)}|^2 \leq \frac{c_{26}}{2^{2    \cdot d
    \cdot l}} .
\end{equation}
\end{lemma}

\noindent
{\bf Proof.}
The result follows by an easy modification of the proof of Lemma 7
in Kohler and Krzy\.zak (2022). The only difference in the proof
is that at the very beginning
we use a sequence of coverings
$\P^{(0)}=\{[0,1]^d\}$, $\P^{(1)}$, \dots, $\P^{(l)}$ of $[0,1]^d$
with the following properties:
\begin{enumerate}
\item
$\P^{(k)}$ consists of $(2^k+1)^d$ many pairwise disjoint cubes of side length $1/2^k$
$(k=1, \dots, l)$.

\item
$[0,1]^d \subseteq \cup_{A \in \P^{(k)}} A$

\item
\begin{equation}
\label{ple6eq1}
\sum_{i=1}^N \PROB_{Z_i} \left(
\cup_{A \in \P^{(k)}} A_{border,\delta}
\right)
\leq
4 d \cdot 2^k \cdot N \cdot \delta,
\end{equation}
where
\begin{eqnarray*}
A_{border,\delta}
&=&
[u^{(1)}-\delta, v^{(1)}+\delta] \times \dots \times
[u^{(d)}-\delta, v^{(d)}+\delta]
\\
&&
\hspace*{3cm}
\setminus
[u^{(1)}+\delta, v^{(1)}-\delta] \times \dots \times
[u^{(d)}+\delta, v^{(d)}-\delta]
\end{eqnarray*}
for
\[
A=[u^{(1)}, v^{(1)}] \times \dots \times
[u^{(d)}, v^{(d)}].
\]
\end{enumerate}

We can ensure (\ref{ple6eq1}) by shifting a partition of
\[
\left[ - \frac{1}{2^k}, 1 \right]^d
\]
consisting of $(2^k+1)^d$ many cubes of side length $1/2^k$
separately
in each component by multiples of $2 \cdot \delta$ less than or
equal to $1/2^k$, which gives us for each component
\[
\left\lfloor \frac{1}{2 \cdot \delta} \cdot \frac{1}{2^k} \right \rfloor
\]
disjoint sets of which at least one must have $\sum_{i=1}^N \PROB_{Z_i}$-measure
less than or equal to
\[
\frac{N}{\lfloor \frac{1}{2 \cdot\delta} \cdot \frac{1}{2^k} \rfloor}
\leq
\frac{N}{ \frac{1}{2 \cdot\delta} \cdot \frac{1}{2^k} -1 }
\leq
N \cdot
\frac{2 \cdot \delta \cdot 2^k}{1-2 \cdot \delta \cdot 2^k} \leq
4 \cdot N \cdot \delta \cdot 2^k
\]
in case $2 \cdot \delta \cdot 2^k \leq 1/2$, which we can assume
w.l.o.g. (because otherwise (\ref{ple6eq1}) is always satisfied).

From this we get the assertion as in the proof of Lemma 7 in Kohler
and Krzy\.zak (2022). \hfill $\Box$

\noindent
{\bf Proof of Lemma \ref{le6a}.}
We apply Lemma \ref{le6} with $d=\kappa^2$ and
$N=(d_1-\kappa+1) \cdot (d_2-\kappa+1)$
and choose
$w_k, w_{t_1,t_2,s_1,s_2,k}^{(l)}, w_{s_2,k}^{(l)}$ such that we have
\[
	o_{(i,j),1,k}^{(L)}
	=f_{\bw,k,1}^{(L)}(\bx_{(i,j)+I})
\quad \mbox{for all } (i,j) \in \{1, \dots, d_1-\kappa+1\}
\times \{1, \dots, d_2-\kappa+1 \}
\]
and
\[
w_k = w_{1,1,k}^{(L)} \quad (k=1, \dots, \tilde{K}_n).
\]
This implies
\begin{eqnarray*}
&&
\int
|
\sum_{k=1}^{\tilde{K}_n}
w_{k} \cdot f_{\bar{\bw}_k,\bar{\bw}_{bias,k}}^{(L)}(\bx)-m(x)|^2 \PROB_{\bX} (d\bx)
\\
&&
\leq
\frac{1}{
(d_1-\kappa+1) \cdot (d_2-\kappa+1)
}
\cdot
\sum_{ i \in \{1, \dots, d_1-\kappa+1\},
  \atop
  j \in \{1, \dots, d_2-\kappa+1\}
		}
\int
\Bigg|
\sum_{k=1}^{\tilde{K}_n} w_{1,1,k}^{(L)} \cdot
f_{\bar{\bw},k,1}^{(L)}(\bx_{(i,j)+I)})
\\
&&
\hspace*{7cm}
-
		f\left(
		\bx_{(i,j)+I}
		\right)
\Bigg|^2 \PROB_{\bX}(d \bx)
\end{eqnarray*}
from which we get the assertion by Lemma \ref{le6}.
\hfill $\Box$

In order to be able to formulate our next auxiliary result we
need the following notation:
Let $(x_1,y_1), \dots, (x_n,y_n) \in \Rd \times \R$, let $K \in \N$,
let $B_1,\dots,B_K:\Rd \rightarrow \R$ and let $c_{27}>0$. In the next lemma
we consider the problem to minimize
\begin{equation}
  \label{se4sub1eq1}
  F(\ba) =
  \frac{1}{n} \sum_{i=1}^n
  |\sum_{k=1}^K a_k \cdot B_k(x_i)-y_i|^2
  +
    c_{27} \cdot  \sum_{k=1}^{K_n} a_k^2 ,
  \end{equation}
where $\ba=(a_1,\dots,a_K)^T$,
by gradient descent. To do this, we choose $\ba^{(0)} \in \R^K$
and set
\begin{equation}
  \label{se4sub1eq2}
  \ba^{(t+1)} = \ba^{(t)}
  - \lambda_n \cdot (\nabla_\ba F)(\ba^{(t)})
    \end{equation}
for some properly chosen $\lambda_n>0.$

        \begin{lemma}
          \label{le7}
          Let $F$ be defined by (\ref{se4sub1eq1}) and choose $\ba_{opt}$
          such that
          \[
F(\ba_{opt})=\min_{\ba \in \R^{K}} F(\ba).
          \]
          Then for any
          $\ba \in \R^{K}$
          we have
                    \[
                            \|(\nabla_\ba F)(\ba)\|^2
                            \geq 4 \cdot c_{27}  \cdot (F(\ba)-F(\ba_{opt})).
                            \]
          \end{lemma}

        \noindent
            {\bf Proof.}
 See Lemma 8  in Drews and Kohler (2022).
    \hfill $\Box$

\subsection{Proof of Theorem \ref{th1}}
\label{se4sub2}
The result follows by a more or less straightforward modification of the proof
of Theorem 1 in Kohler and Krzy\.zak (2022) using Lemma \ref{le5a}
and Lemma \ref{le6a} instead of Lemma \ref{le5} and Lemma \ref{le6}.
For the sake of completeness we nevertheless present a complete proof.

For $z \in \R$ we have that
$T_1 z > 1/2$ holds if and only if $z>1/2$,
hence we can assume w.l.o.g. that our estimate
is given by
\[
f_n(\bx)=
\begin{cases}
	1, & \mbox{if } m_n(\bx) \geq \frac{1}{2}, \\
	0, & \mbox{elsewhere}
\end{cases}
\]
where
\[
m_n(\bx)= T_{1} \left( f_{\bw^{(t_n)}}(\bx) \right).
\]
Consequently we know by Lemma \ref{le1} that it suffices to show
that we have for any $\epsilon>0$
\begin{equation}
\label{pth1eq1}
\EXP \int |m_n(\bx)-\eta(\bx)|^2 \PROB_X(d \bx)
\leq
c_{28} \cdot n^{- \frac{1}{\kappa^2 +1}+\epsilon}
\end{equation}

Set $I=\{1, \dots, \kappa\} \times \{1, \dots, \kappa\}$,
$r=(d_1-\kappa+1) \cdot (d_2-\kappa+1)$, $\delta=c_{29} \cdot n^{-1/(1+
  \cdot \kappa^2)}$ and
\[
\tilde{K}_n = n^6.
\]
Using Lemma \ref{le6a}
(with $\delta=n^{-1/(1+\kappa^2)}$ and sufficiently large $s$)
we can construct a weight vector $\bw$ of
a convolutional neural network
\[
\tilde{f}_{\bw}(\bx)=\sum_{k=1}^{\tilde{K}_n} w_k \cdot f_{\bw_k,\bw_{bias,k}}(\bx)
\]
with the property that for any weight vector $\bar{\bw}$ with
\[
|\bar{w}_{t_1,t_2,s_1,s_2,k}^{(r)} - w_{t_1,t_2,s_1,s_2,k}^{(r)}| \leq \log n
\quad \mbox{and} \quad
|\bar{w}_{s_2,k}^{(r)}-w_{s_2,k}^{(r)}| \leq \log n
\]
$(r=1, \dots, L)$
we have
\begin{eqnarray}
&&
\int
\left|
\sum_{k=1}^{\tilde{K}_n}
w_k \cdot
f_{
\bar{\bw}_k, \bar{\bw}_{bias,k}
}(\bx)
-
\frac{
\sum_{(i,j) \in \{1, \dots, d_1-\kappa+1\} \times \{1, \dots,
   d_2-\kappa+1\}}
f(\bx_{(i,j)+I})
}{(d_1-\kappa+1) \cdot (d_2-\kappa+1)}
\right|^2
\PROB_{\bX}(d \bx)
\nonumber \\
&&
\leq
c_{30} \cdot (\log n)^2 \cdot n^{- \frac{1}{1+\kappa^2}}
\label{pth1eq*}
\end{eqnarray}
and
\[
\sum_{k=1}^{\tilde{K}_n} w_k^2 \leq \frac{c_{31}}{n}.
\]
Let $A_n$ be the event that there exists pairwise distinct
$j_1, \dots, j_{\tilde{K}_n} \in \{1, \dots, K_n\}$ such that the
randomly initialized weights satisfy
\begin{equation}
\label{pth1eq2}
|(\bw^{(0)})_{t_1,t_2,s_1,s_2,j_k}^{(r)} - w_{t_1,t_2,s_1,s_2,k}^{(r)}| \leq \log n
\end{equation}
and
\begin{equation}
\label{pth1eq3}
|(\bw^{(0)})_{s_2,j_k}^{(r)}-w_{s_2,k}^{(r)}| \leq \log n
\end{equation}
for $r=1, \dots, L$ and $k=1, \dots, \tilde{K}_n$.

Define the weight vectors $(\bw^*)^{(t)}$ $(t=0,1, \dots, t_n)$ by
\[
((\bw^*)^{(t)})_{j_k} = w_k \quad \mbox{if } k \in \{1, \dots, \tilde{K}_n\},
\]
\[
((\bw^*)^{(t)})_{k} = 0 \quad \mbox{if } k \notin \{j_1, \dots, j_{\tilde{K}_n}\},
\]
\[
((\bw^*)^{(t)})_{s_1,s_2,t_1,t_2,k}^{(l)}=
(\bw^{(t)})_{s_1,s_2,t_1,t_2,k}^{(l)}
\quad \mbox{if }
l \in \{1, \dots, L\}
\]
and
\[
((\bw^*)^{(t)})_{s_2,k}^{(l)}=
(\bw^{(t)})_{s_2,k}^{(l)}
\quad \mbox{if }
l \in \{1, \dots, L\}.
\]
In order to show (\ref{pth1eq1}) we will use the following error decomposition:
\begin{eqnarray*}
&&
\int | m_n(\bx)-\eta(\bx)|^2 \PROB_{\bX} (d \bx)
\\
&&
=
\left(
\EXP \left\{ |m_n(\bX)-Y|^2 | \D_n \right\}
-
\EXP \{ |\eta(\bX)-Y|^2\}
\right)
\cdot 1_{A_n}
+
\int | m_n(\bx)-\eta(\bx)|^2 \PROB_{\bX} (d \bx)
\cdot 1_{A_n^c}
\\
&&
=
\Big[
\EXP \left\{ |m_n(\bX)-Y|^2 | \D_n \right\}
-
\EXP \{ |\eta(\bX)-Y|^2\}
\\
&&
\hspace*{2cm}
- \left(
2 \cdot \frac{1}{n} \sum_{i=1}^n
|m_n(\bX_i)-Y_i|^2
-
2 \cdot \frac{1}{n} \sum_{i=1}^n
|\eta(\bX_i)- Y_i|^2
\right)
\Big] \cdot 1_{A_n}
\\
&&
\quad
+
\Big[
2 \cdot \frac{1}{n} \sum_{i=1}^n
|m_n(\bX_i)-Y_i|^2
-
2 \cdot \frac{1}{n} \sum_{i=1}^n
|\eta(\bX_i)- Y_i|^2
\Big] \cdot 1_{A_n}
\\
&&
\quad
+
\int | m_n(\bx)-\eta(\bx)|^2 \PROB_{\bX} (dx)
\cdot 1_{A_n^c}
\\
&&
=: \sum_{j=1}^3 T_{j,n}.
\end{eqnarray*}
In the reminder of the proof we bound
\[
\EXP T_{j,n}
\]
for $j \in \{1, 2, 3\}$.

In the {\it first step of the proof} we show
\[
\EXP T_{3,n} \leq \frac{c_{32}}{n}.
\]
The definition of $m_n$ implies $\int |m_n(\bx)-\eta(\bx)|^2 \PROB_{\bX}
(d \bx) \leq
4$, hence it suffices to show
\begin{equation}
\label{pth1eq4}
\PROB(A_n^c) \leq \frac{c_{33}}{n}.
\end{equation}
To do this, we consider sequential choice of the initial weights of the
$K_n$ convolutional neural networks which we compute in parallel.

 Probability that the weights in
the first of these networks differ in all components by at most $\log n$
from $w_{t_1,t_2,s_1,s_2,1}^{(l)}$, $w_{s_2,1}^{(l)}$$(l=1, \dots, L)$ is
for large $n$ bounded below by
\begin{eqnarray*}
 \left( \frac{\log n}{2 \cdot c_2 \cdot (\log n)^2}
\right)^{2 \cdot \kappa^2 \cdot (2 \cdot \kappa^2+1) \cdot (L-1)}
\cdot
\left(
\frac{\log n}{2 \cdot c_3 \cdot n^\tau}
\right)^{2 \cdot \kappa^2 \cdot (\kappa^2+1)}
&\geq&
 n^{- 2 \cdot \kappa^2 \cdot (\kappa^2+1) \cdot \tau-0.5}.
\end{eqnarray*}
Hence probability that none of the first $n^{2 \cdot \kappa^2 \cdot (\kappa^2+1) \cdot
  \tau +1}$ neural networks satisfies this condition is for large $n$
 bounded above by
\begin{eqnarray*}
(1 -  n^{- 2 \cdot \kappa^2 \cdot (\kappa^2+1) \cdot \tau-0.5}) ^{n^{2 \cdot \kappa^2 \cdot (\kappa^2+1) \cdot
  \tau +1}}
&\leq&
\left(\exp \left(
-  n^{- 2 \cdot \kappa^2 \cdot (\kappa^2+1) \cdot \tau-0.5}
\right)
\right) ^{n^{2 \cdot \kappa^2 \cdot (\kappa^2+1) \cdot
  \tau +1}}
\\
&=&
\exp( -  n^{0.5}).
\end{eqnarray*}
Since we have $K_n \geq n^{2 \cdot \kappa^2 \cdot (\kappa^2+1) \cdot
  \tau +1} \cdot \tilde{K}_n$ for large  $n$ we can successively
use the same construction for all of $\tilde{K}_n$ weights and we can conclude:
Probability that there exists $k \in \{1, \dots, \tilde{K}_n\}$
such that
none of the $K_n$ weight vectors of the convolutional
neural network differs by at most $\log n$ from
$w_{t_1,t_2,s_1,s_2,k}^{(l)}$, $w_{s_2,k}^{(l)}$ is for large $n$ bounded from above by
\begin{eqnarray*}
&&
\PROB(A_n^c)=
\tilde{K}_n \cdot \exp( -  n^{0.5})
\leq  n^\rho \cdot \exp( -  n^{0.5})
\leq \frac{c_{33}}{n}.
\end{eqnarray*}

In the {\it second step of the proof} we show for large  $n$
\begin{equation}
\label{pth1eq5}
\| \bw^{(t)}-\bw^{(0)}\| \leq \log n
\end{equation}
for all $t=1, \dots, t_n$.
For large  $n$ we have
\[
F_n(\bw^{(0)}) = \frac{1}{n} \sum_{i=1}^n |Y_i-0|^2 + 0 \leq 1
\]
and
\[
2 \cdot \frac{t_n}{L_n} \leq (\log n)^2.
\]
Application of Lemma \ref{le3}
and Lemma \ref{le4}
with
$\gamma_n^*=\log n$
and
$B_n=(c_2 + 1) \cdot (\log n)^2$
yields that the assumptions
(\ref{le2eq1}) and (\ref{le2eq2}) of Lemma \ref{le2} are satisfied.
Lemma \ref{le2} implies the assertion.

Let $\epsilon >0$ be arbitrary.
In the {\it third step of the proof} we show
\[
\EXP T_{1,n} \leq
c_{34} \cdot
\frac{ n^{\tau \cdot \kappa^2 + \epsilon}}{n}
.
\]
Let $\W_n$ be the set of all weight vectors
$\bw$ which satisfy
\[
| w_k| \leq  (\log n)^2 \quad (k=1, \dots, K_n),
\]
\[
\max\left\{
|w_{t_1,t_2,s_1,s_2,k}^{(l)}|, |w_{s_2,k}^{(l)}|
\right\}
\leq (c_2+1) \cdot (\log n)^2 \quad (l=2, \dots, L)
\]
and
\[
\max\left\{
|w_{t_1,t_2,s_1,s_2,k}^{(1)}|, |w_{s_2,k}^{(1)}|
\right\}
 \leq (c_3+1) \cdot (\log n)^2 \cdot n^\tau.
\]
By the second step and
the initial choice of $\bw^{(0)}$ we know that on
$A_n$
we have
\[
\bw^{(t)} \in \W_n \quad (t=0, \dots, t_n).
\]
Hence, for any $u>0$ we get
for large  $n$
\begin{eqnarray*}
&&
\PROB \{ T_{1,n} > u \}
\\
&&
\leq
\PROB \Bigg\{
\exists f \in \F_n :
\EXP \left(
\left|
f(\bX) - Y
\right|^2
\right)
-
\EXP \left(
\left|
\eta(\bX)-Y
\right|^2
\right)
\\
&&\hspace*{3cm}-
\frac{1}{n} \sum_{i=1}^n
\left(
\left|
f(\bX_i) - Y_i
\right|^2
-
\left|
\eta(\bX_i)-Y_i
\right|^2
\right)
\\
&&\hspace*{2cm}
> \frac{1}{2} \cdot
\left(
u
+
\EXP \left(
\left|
f(\bX)-Y
\right|^2
\right)
-
\EXP \left(
\left|
\eta(\bX)-Y
\right|^2
\right)
\right) \Bigg\}
,
\end{eqnarray*}
where
\[
\F_n = \left\{ T_{1} (f_\bw) \quad : \quad \bw \in \W_n \right\}.
\]
By Lemma \ref{le5a} we have
\begin{eqnarray*}
&&
\Nu_1 \left(
\delta, \F_n
, x_1^n
\right)
\leq
\left(
\frac{ c_{34}}{\delta}
\right)^{
c_{35} \cdot (\log n)^{2 \kappa^2}  n^{\tau \cdot \kappa^2} \cdot (\log n)^{2 \cdot (L-1) \cdot \kappa^2}
\cdot
   \left(\frac{K_n \cdot (\log n)^2}{\delta}\right)^{\kappa^2/k} + c_{36}
  }.
\end{eqnarray*}
By choosing $k$ large enough we get for $\delta>1/n^2$
\[
\Nu_1 \left(
\delta , \F_n
, x_1^n
\right)
\leq
c_{37} \cdot n^{ c_{38} \cdot n^{\tau \cdot \kappa^2  + \epsilon/2}}.
\]
This together with Theorem 11.4 in Gy\"orfi et al. (2002) leads for $u
\geq 1/n$ to
\[
\PROB\{T_{1,n}>u\}
\leq
14 \cdot
c_{37} \cdot n^{ c_{38} \cdot n^{\tau \cdot \kappa^2 +  \epsilon/2}}
\cdot
\exp \left(
- \frac{n}{5136} \cdot u
\right).
\]
For $\epsilon_n \geq 1/n$ we can conclude for large  $n$
\begin{eqnarray*}
\EXP \{ T_{1,n} \}
& \leq &
\epsilon_n + \int_{\epsilon_n}^\infty \PROB\{ T_{1,n}>u \} \, du
\\
& \leq &
\epsilon_n
+
14 \cdot
c_{37} \cdot n^{ c_{38} \cdot n^{\tau \cdot \kappa^2 +
    \epsilon/2}}
\cdot
\exp \left(
- \frac{n}{5136} \cdot \epsilon_n
\right)
\cdot
\frac{5136}{n}.
\end{eqnarray*}
Setting
\[
\epsilon_n = \frac{5136}{n}
\cdot
c_{38}
\cdot
 n^{\tau \cdot \kappa^2 +
    \epsilon/2}
\cdot \log n
\]
yields the assertion of the third step of the proof.

In the {\it fourth step of the proof} we show
\begin{eqnarray*}
&&
  \EXP \{ T_{2,n} \} \leq
  c_{39} \cdot
  \left(
\frac{n^{\tau \cdot \kappa^2 + \epsilon}}{n} + n^{- \frac{1}{1+\kappa^2}}
  \right).
\end{eqnarray*}
Using
\[
|T_{1}( z) - y| \leq |z-y|
\quad \mbox{for } |y| \leq 1
\]
we get
\begin{eqnarray*}
&&
T_{2,n}/2
\\
&&
=
\Big[ \frac{1}{n} \sum_{i=1}^n
|m_n(\bX_i)-Y_i|^2
-
 \frac{1}{n} \sum_{i=1}^n
|\eta(\bX_i)- Y_i|^2
\Big] \cdot 1_{A_n}
\\
&&
\leq
\Big[
\frac{1}{n} \sum_{i=1}^n
|f_{\bw^{(t_n)}}(\bX_i)-Y_i|^2
-
 \frac{1}{n} \sum_{i=1}^n
|\eta(\bX_i)- Y_i|^2
\Big] \cdot 1_{A_n}
\\
&&
\leq
\Big[ F_n(\bw^{(t_n)})
-
 \frac{1}{n} \sum_{i=1}^n
|\eta(\bX_i)- Y_i|^2
\Big] \cdot 1_{A_n}.
\end{eqnarray*}
Application of Lemma \ref{le2} (which is possible due to
 Lemma \ref{le3} and Lemma \ref{le4}) implies that this in turn
is less than
\begin{eqnarray*}
&&
\Big[
F_n(\bw^{(t_n-1)}) - \frac{1}{2 L_n} \cdot \| \nabla_\bw F_n(\bw^{(t_n-1)}) \|^2
-
 \frac{1}{n} \sum_{i=1}^n
|\eta(X_i)- Y_i|^2
\Big] \cdot 1_{A_n}.
\end{eqnarray*}
Since the sum of squares of all partial derivatives is at least
as large as the sum of squares of the partial derivatives with respect
to the outer weights $w_k$ $(k=1, \dots, K_n)$, we can
upper bound this in turn following Lemma \ref{le7} by
\begin{eqnarray*}
&&
\Big[
F_n(\bw^{(t_n-1)}) - \frac{1}{2 L_n} \cdot 4 \cdot c_4 \cdot
  ( F_n(\bw^{(t_n-1)}) -  F_n((\bw^*)^{(t_n-1)})
\\
&&
\hspace*{6cm}
-
 \frac{1}{n} \sum_{i=1}^n
|\eta(\bX_i)- Y_i|^2
\Big] \cdot 1_{A_n}
\\
&&
=
\Big[
 \left(
1 - \frac{2 \cdot c_4}{ L_n}
\right)
\cdot
F_n(\bw^{(t_n-1)})
+
\frac{2 \cdot c_4}{ L_n} \cdot
F_n((\bw^*)^{(t_n-1)})
-
 \frac{1}{n} \sum_{i=1}^n
|\eta(\bX_i)- Y_i|^2
\Big] \cdot 1_{A_n}.
\end{eqnarray*}
Applying this argument repeatedly shows that
\begin{eqnarray*}
&&
T_{2,n}/2
\\
&&
\leq
\Big[
 \left(
1 - \frac{2 \cdot c_4}{ L_n}
\right)^{t_n}
\cdot
F_n(\bw^{(0)})
+
\sum_{k=1}^{t_n}
\frac{2 \cdot c_4}{ L_n} \cdot
  \left(
1 - \frac{2 \cdot c_4}{ L_n}
\right)^{k-1}
F_n((\bw^*)^{(t_n-k)})
\\
&&
\hspace*{3cm}
-
 \frac{1}{n} \sum_{i=1}^n
|\eta(\bX_i)- Y_i|^2
\Big] \cdot 1_{A_n}.
\end{eqnarray*}
This implies
\begin{eqnarray*}
&&
\EXP\{ T_{2,n}/2 \}
\\
&&
\leq
\left(
1 - \frac{2 \cdot c_4}{ L_n}
\right)^{t_n}
\cdot \EXP \{Y^2\}
+
\sum_{k=1}^{t_n}
\frac{2 \cdot c_4}{ L_n} \cdot
  \left(
1 - \frac{2 \cdot c_4}{ L_n}
\right)^{k-1}
\cdot
\\
&&
\hspace*{0.5cm}
\EXP \Bigg(
\Bigg(
\frac{1}{n} \sum_{i=1}^n
|
f_{(\bw^*)^{(t_n-k)}} (\bX_i)-Y_i|^2
-
\frac{1}{n} \sum_{i=1}^n
|\eta(\bX_i)-Y_i|^2
\Bigg) \cdot
   1_{A_n} \Bigg)
\\
&&
\quad
+ c_4 \cdot \sum_{k=1}^{\tilde{K}_n} |w_k |^2
\\
&&
\leq
\left(
1 - \frac{2 \cdot c_4}{ L_n}
\right)^{t_n}
\cdot \EXP \{Y^2\} + c_4 \cdot \sum_{k=1}^{\tilde{K}_n} |w_k|^2
\\
&&
\quad
+
\sum_{k=1}^{t_n}
\frac{2 \cdot c_4}{ L_n} \cdot
  \left(
1 - \frac{2 \cdot c_4}{ L_n}
\right)^{k-1}
\cdot 2 \cdot
\\
&&
\hspace*{1cm}
\EXP \Bigg(
\max_{
k=0, \dots, t_n-1
}
\int
|
f_{(\bw^*)^{(t_n-k)}} (\bx)-\eta(x)|^2 \PROB_X (d \bx)
\Bigg)
\\
&&
\quad
+
\sum_{k=1}^{t_n}
\frac{2 \cdot c_4}{ L_n} \cdot
  \left(
1 - \frac{2 \cdot c_4}{ L_n}
\right)^{k-1}
\cdot
\\
&&
\hspace*{0.5cm}
\EXP \Bigg(
\Bigg(
\frac{1}{n} \sum_{i=1}^n
|
f_{(\bw^*)^{(t_n-k)}} (\bX_i)-Y_i|^2
-
\frac{1}{n} \sum_{i=1}^n
|\eta(\bX_i)-Y_i|^2
\\
&&
\hspace*{0.5cm}
- 2 \cdot \Bigg(
 \EXP\{ |
f_{(\bw^*)^{(t_n-k)}} (\bX)-Y|^2 | \D_n,
   \bw^{(0)}\}
\\
&&
\hspace*{4cm}
-\EXP\{|\eta(\bX)-Y|^2\}
\Bigg)
\Bigg) \cdot
   1_{A_n} \Bigg).
\end{eqnarray*}
Arguing as in the third step of the proof
(which is possible even if we do not have truncated functions because
of (\ref{pth1eq5}) and (\ref{le6aeq1}))
we get
\begin{eqnarray*}
&&
\EXP \Bigg(
\Bigg(
\frac{1}{n} \sum_{i=1}^n
|
f_{(\bw^*)^{(t_n-k)}} (\bX_i)-Y_i|^2
-
\frac{1}{n} \sum_{i=1}^n
|\eta(\bX_i)-Y_i|^2
\\
&&
\hspace*{0.5cm}
- 2 \cdot \Bigg(
 \EXP\{ |
f_{(\bw^*)^{(t_n-k)}} (\bX)-Y|^2 | \D_n,
   \bw^{(0)}\}
\\
&&
\hspace*{4cm}
-\EXP\{|\eta(\bX)-Y|^2\}
\Bigg)
\Bigg) \cdot
   1_{A_n}
\Bigg)
\\
&&
\leq
c_{40} \cdot \frac{(\log n)^2}{n}
+
c_{41} \cdot
\frac{ n^{\tau \cdot \kappa^2 + \epsilon}}{n}.
\end{eqnarray*}

From this we
conclude
\begin{eqnarray*}
&&
\EXP\{ T_{2,n}/2 \} \\
&&
\leq
\left(
1 - \frac{2 \cdot c_4}{ L_n}
\right)^{t_n}
\cdot \EXP \{Y^2\}
\\
&&
\quad
+
2 \cdot
\EXP \Bigg(
\max_{
k=0, \dots, t_n-1
}
\int
|
\tilde{f}_{(\bw^*)^{(t_n-k)}} (\bx)-\eta(x)|^2 \PROB_X (d \bx)
\Bigg)
\\
&&
\quad
+ c_4 \cdot \sum_{k=1}^{\tilde{K}_n} |w_k|^2+
c_{40} \cdot \frac{(\log n)^2}{n}
+
c_{41} \cdot
\frac{ n^{\tau \cdot \kappa^2 + \epsilon}}{n}.
\end{eqnarray*}
The definition of $t_n$
together with $c_5 \geq 1/(2 \cdot c_4)$ implies
\begin{eqnarray*}
\left(
1 - \frac{2 \cdot c_4}{ L_n}
\right)^{t_n}
\cdot \EXP \{Y^2\}
&\leq&
\exp \left(
-
\frac{2 \cdot c_4}{ L_n} \cdot t_n
\right)
\cdot \EXP \{Y^2\}
\\
&\leq&
\exp ( -2 \cdot c_4 \cdot c_5 \cdot \log n) \cdot \EXP \{Y^2\}
\\
&\leq& \frac{c_{42}}{n}.
\end{eqnarray*}
And by (\ref{pth1eq*}) we know
\[
\max_{
k=0, \dots, t_n-1
}
\int
|
\tilde{f}_{(\bw^*)^{(t_n-k)}} (\bx)-\eta(x)|^2 \PROB_X (d \bx)
\leq
c_{43} \cdot (\log n)^2 \cdot n^{- \frac{1}{1+\kappa^2}}
\]
All the results above imply the assertion.

\hfill $\Box$

\section{Acknowledgment}
The first and the third author  would like to thank
the Deutsche Forschungsgemeinschaft (DFG, German Research  Foundation)
for funding this project (Projektnummer 449102119). The second author would like to thank
Natural Sciences and Engineering Research Council of Canada for funding this project under Grant RGPIN-2020-06793.

\end{document}